\documentclass{article}

\usepackage{amsmath,amsfonts,bm}

\newcommand{\T}{\top}
\newcommand{\A}{\textrm{A}}
\newcommand{\B}{\textrm{B}}

\newcommand{\tot}{\textrm{tot}}

\newcommand{\lnorm}{\left \|}
\newcommand{\rnorm}{\right \|}

\def\eqref#1{equation~\ref{#1}}

\def\1{\bm{1}}

\def\vzero{{\bm{0}}}
\def\vone{{\bm{1}}}

\def\ve{{\bm{e}}}

\def\vr{{\bm{r}}}

\def\vw{{\bm{w}}}
\def\vx{{\bm{x}}}

\def\mI{{\bm{I}}}

\def\mW{{\bm{W}}}

\def\mSigma{{\bm{\Sigma}}}

\DeclareMathAlphabet{\mathsfit}{\encodingdefault}{\sfdefault}{m}{sl}
\SetMathAlphabet{\mathsfit}{bold}{\encodingdefault}{\sfdefault}{bx}{n}

\def\gM{{\mathcal{M}}}

\def\sR{{\mathbb{R}}}

\newcommand{\Ls}{\mathcal{L}}

\newcommand{\Var}{\mathrm{Var}}

\usepackage{microtype}
\usepackage{graphicx}
\usepackage{subcaption}
\usepackage{booktabs} 
\usepackage{dblfloatfix}

\usepackage{hyperref}
\usepackage{xurl}

\usepackage[accepted]{icml2024}

\usepackage{amsmath}
\usepackage{amssymb}
\usepackage{mathtools}
\usepackage{amsthm}

\usepackage[capitalize,noabbrev]{cleveref}

\theoremstyle{plain}

\theoremstyle{definnetworkition}

\theoremstyle{remark}

\usepackage[textsize=tiny]{todonotes}

\icmltitlerunning{Understanding Unimodal Bias in Multimodal Deep Linear Networks}

\begin{document}

\twocolumn[
\icmltitle{Understanding Unimodal Bias in Multimodal Deep Linear Networks}




\begin{icmlauthorlist}
\icmlauthor{Yedi Zhang}{gatsby}
\icmlauthor{Peter E. Latham}{gatsby}
\icmlauthor{Andrew Saxe}{gatsby,swc}
\end{icmlauthorlist}

\icmlaffiliation{gatsby}{Gatsby Computational Neuroscience Unit, University College London}
\icmlaffiliation{swc}{Sainsbury Wellcome Centre, University College London}

\icmlcorrespondingauthor{Yedi Zhang}{yedi@gatsby.ucl.ac.uk}

\icmlkeywords{deep linear networks, multimodal learning, unimodal bias}

\vskip 0.3in
]



\printAffiliationsAndNotice{}  

\begin{abstract}
Using multiple input streams simultaneously to train multimodal neural networks is intuitively advantageous but practically challenging. A key challenge is unimodal bias, where a network overly relies on one modality and ignores others during joint training. We develop a theory of unimodal bias with multimodal deep linear networks to understand how architecture and data statistics influence this bias. This is the first work to calculate the duration of the unimodal phase in learning as a function of the depth at which modalities are fused within the network, dataset statistics, and initialization. We show that the deeper the layer at which fusion occurs, the longer the unimodal phase. A long unimodal phase can lead to a generalization deficit and permanent unimodal bias in the overparametrized regime.
Our results, derived for multimodal linear networks, extend to nonlinear networks in certain settings.
Taken together, this work illuminates pathologies of multimodal learning under joint training, showing that late and intermediate fusion architectures can give rise to long unimodal phases and permanent unimodal bias.
Our code is available at: https://yedizhang.github.io/unimodal-bias.html.
\end{abstract}

\section{Introduction}
The success of multimodal deep learning hinges on effectively utilizing multiple modalities \citep{morency19survey,morency22survey}. 
However, some multimodal networks overly rely on a faster-to-learn or easier-to-learn modality and ignore the others during joint training \citep{goyal17cvpr,cadene19RUBi,wang20cvpr,gat21score,peng22cvpr}. 
For example, Visual Question Answering models should provide a correct answer by both ``listening'' to the question and ``looking'' at the image \citep{agrawal16analyzing}, whereas they tend to overly rely on the language modality and ignore the visual modality \citep{goyal17cvpr,agrawal18cvpr,hessel20emnlp}.
This phenomenon has been observed in a variety of settings, and has several names: unimodal bias \citep{cadene19RUBi}, greedy learning \citep{wu22greedy}, modality competition \citep{huang22competition}, modality laziness \citep{du23laziness}, and modality underutilization \citep{nyu23latent}.
In this paper, we adopt the term unimodal bias to refer to the phenomenon in which a multimodal network learns from different input modalities at different times during joint training.  

The extent to which multimodal networks exhibit unimodal bias depends on both the dataset and the multimodal network architecture. Regarding datasets, practitioners managed to alleviate the bias by building more balanced multimodal datasets \citep{goyal17cvpr,agrawal18cvpr,hudson19GQA}.
Regarding multimodal network architectures, empirical work has shown that unimodal bias emerges in jointly trained late fusion networks \citep{wang20cvpr,huang22competition} and intermediate fusion networks \citep{wu22greedy}, while early fusion networks may encourage usage of all input modalities \citep{gadzicki20earlyVSlate,barnum20early}.

Despite empirical evidence, there is scarce theoretical understanding of how unimodal bias is affected by the network configuration, dataset statistics, and initialization.
To fill the gap, we study multimodal deep linear networks where unimodal bias manifests in ways consistent with those observed in complex networks, thereby serving as a prerequisite for theoretical understanding of complex networks.
We show that unimodal bias is conspicuous in late and intermediate fusion linear networks but not in early fusion linear networks. Intermediate and late fusion linear networks learn one modality first and the other after a delay, yielding a phase in which the multimodal network implements a unimodal function. 
We calculate the duration of the unimodal phase in terms of parameters of the network and statistics of the dataset. We find that a deeper fusion layer within the multimodal network, stronger correlations between input modalities, and greater disparities in input-output correlations for each modality all prolong the unimodal phase.
In the overparameterized regime, long unimodal phases creates a dilemma between overfitting one modality and underfitting the other, resulting in permanent unimodal bias and a generalization deficit.
Our results apply to nonlinear networks in certain settings, providing insights for examining, diagnosing, and curing unimodal bias in a broader range of realistic cases.

In summary, our contributions are the following: 
(i) We provide a theoretical explanation for why unimodal bias is conspicuous in late and intermediate fusion linear networks but not in early fusion linear networks. 
(ii) We calculate the duration of the unimodal phase in mulitmodal learning with late and intermediate fusion linear networks, as a function of the network configuration, correlation matrices of the dataset, and initialization scale. 
(iii) We analyze the mis-attribution during the unimodal phase and the superficial modality preference.
(iv) We reveal that long unimodal phases lead to a generalization deficit and permanent unimodal bias in the overparameterized regime.
(v) We validate our findings with numerical simulations of multimodal deep linear networks and certain nonlinear networks.

\subsection{Related Work}
Several attempts have been made to understand unimodal bias.
On the empirical front, \citet{wang20cvpr,gat21score,wu22greedy,peng22cvpr,kleinman23critical} have proposed metrics to assess unimodal bias across various multimodal models. Their metrics contributed to multimodal learning applications but did not lend a theory.
On the theoretical front, \citet{huang21better,huang22competition} proved that multimodal learning can outperform unimodal learning but may fail to deliver due to modality competition. 
In such failed cases, multiple modalities compete; only a subset of modalities correlated more with their encoding network's random initialization will win and be learned by the multimodal network.
However, their study was restricted to two-layer late fusion networks and did not cover the influence of different networks and datasets upon the unimodal bias.
Our work studies multimodal deep linear networks with three different fusion schemes, and we analytically reveal the relationship between unimodal bias, network configuration, and dataset statistics. 

Our work builds on a rich line of theoretical analyses of deep linear networks. 
\citet{fukumizu98batch,saxe13exact,saxe19semantic,lampinen2018gen,clem22prior,shi22pathways,cengiz22silent} obtained exact solutions to the gradient descent dynamics of deep linear networks with whitened input.
\citet{gidel19reg,advani20highd,tarmoun21overparam} derived reductions and a special-case solution assuming spectral initialization for the same dynamics but with unwhitened input. 
Balancing properties in linear networks were discovered and proved in \citep{fukumizu98batch,arora18acc,du18autobalance,ji18align}.
Nonetheless, previous solutions do not directly address our problem because multimodal deep linear networks are not fully connected and multimodal datasets generally do not have whitened input. 
We thus develop new analytic tools for multimodal deep linear networks, extending the theoretical landscape of linear networks.

\section{Problem Setup}

\begin{figure}
\centering
\includegraphics[width=0.8\linewidth]{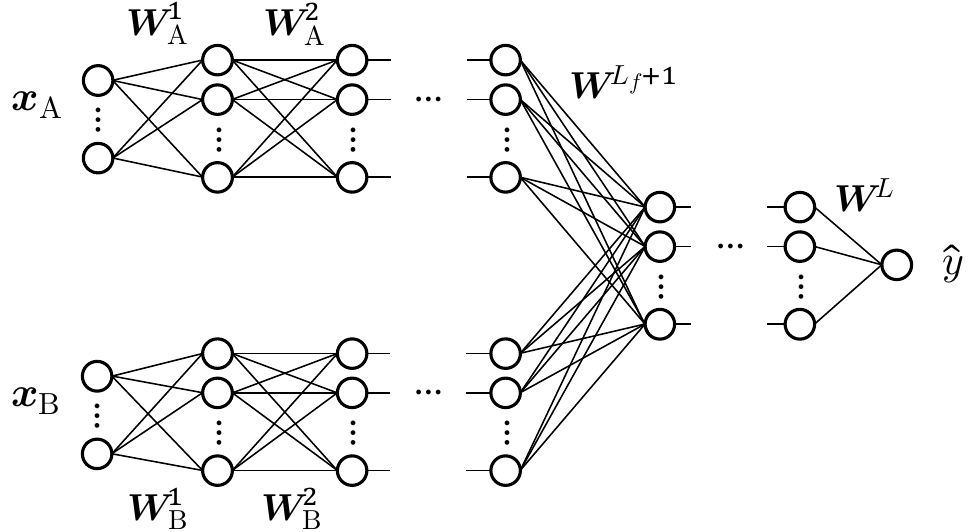}
\caption{Schematic of a multimodal fusion network with total depth $L$ and fusion layer at $L_f$.}
\label{fig:schematic}
\end{figure}

\subsection{Multimodal Data}
Let $\vx \in \sR^D$ represent an arbitrary multimodal input and $y\in \sR$ be its scalar target output. We are given a dataset $\{ \vx^\mu, y^\mu \}_{\mu=1}^P$ consisting of $P$ samples. For simplicity, we assume there are two modalities, A and B, with full input $\vx = [\vx_\A, \vx_\B]$. Since we study multimodal linear networks with mean square error loss, the learning dynamics depends only on the correlation matrices of the data \citep{fukumizu98batch,saxe13exact}. We denote the input correlation matrix as $\mSigma$ and input-output correlation matrix as $\mSigma_{y\vx}$; these are given explicitly by
\begin{subequations}  \label{eq:cov-def}
\begin{align}
\mSigma &= \begin{bmatrix}
\mSigma_\A & \mSigma_{\A \B}  \\
\mSigma_{\B \A} & \mSigma_\B  
\end{bmatrix}
= \begin{bmatrix}
\left\langle \vx_\A \vx_\A^\T \right\rangle & \left\langle \vx_\A \vx_\B^\T \right\rangle  \\
\left\langle \vx_\B \vx_\A^\T \right\rangle & \left\langle \vx_\B \vx_\B^\T \right\rangle
\end{bmatrix}
, \\
\mSigma_{y\vx} &= \begin{bmatrix}
\mSigma_{y\vx_\A}  &  \mSigma_{y\vx_\B}
\end{bmatrix}
= \begin{bmatrix}
\left\langle y \vx_\A^\T \right\rangle  &  \left\langle y \vx_\B^\T \right\rangle
\end{bmatrix} ,
\end{align}
\end{subequations}
where $\langle \cdot \rangle$ denotes the average over the dataset. We assume data points are centered to have zero mean, $\langle \vx \rangle = \vzero$, and the input correlation matrix $\mSigma$ has full rank, but make no further assumptions about the dataset.

\subsection{Multimodal Fusion Linear Network}
We study a multimodal deep linear network with total depth $L$ and fusion layer at $L_f$ defined as\footnote{We abuse the notation $\prod_j \mW^j$ to represent the ordered product of matrices with the largest index on the left and smallest on the right.}
\begin{align}
\hat y (\vx; \mW) 
&= \prod ^{L}_{j=L_f+1} \mW^j \left( \prod ^{L_f}_{i=1} \mW^i_\A \vx_\A + \prod ^{L_f}_{i=1} \mW^i_\B \vx_\B \right)  \nonumber \\
&\equiv \mW^{\tot}_\A \vx_\A + \mW^{\tot}_\B \vx_\B  \nonumber \\
&\equiv \mW^{\tot} \vx \, .
\end{align}
The overall network input-output map is denoted $\mW^{\tot}$, and the map for each modality is denoted $\mW^{\tot}_\A, \mW^{\tot}_\B$. We use $\mW$ to denote all weight parameters collectively. We assume the number of neurons in both pre-fusion layer branches is of the same order. A schematic of this network is given in \cref{fig:schematic}.

Our network definition incorporates bimodal deep linear networks of three common fusion schemes. Categorized by the multimodal deep learning community \citep{rama17survey}, the case where $L_f=1$ is early or data-level fusion; $1<L_f<L$ is intermediate fusion; and $L_f=L$ is late or decision-level fusion.

\subsection{Gradient Descent Dynamics}
The network is trained by optimizing the mean square error, $\Ls = \frac1{2P} \sum_{\mu=1}^P (y^\mu - \hat y^\mu)^2$, with gradient descent. In the limit of small learning rate, the gradient descent dynamics are well approximated by the continuous time differential equations; see \cref{supp:grad}. For pre-fusion layers $1 \leq l \leq L_f$,
\begin{subequations}
\begin{align}
\tau \dot \mW_\A^l &= \left( \prod_{j=L_f+1}^L \mW^j \prod_{i=l+1}^{L_f} \mW^i_\A \right)^\T 
\ve_\A
\left( \prod_{i=1}^{l-1} \mW^i_\A \right)^\T 
, \\
\tau \dot \mW_\B^l &= \left( \prod_{j=L_f+1}^L \mW^j \prod_{i=l+1}^{L_f} \mW^i_\B \right)^\T 
\ve_\B
\left( \prod_{i=1}^{l-1} \mW^i_\B \right)^\T 
.  \label{eq:dWB}
\end{align} \label{eq:dWAdWB}%
\end{subequations}
For post-fusion layers $L_f+1 \leq l \leq L$,
\begin{align}
\tau \dot \mW^l &= \left( \prod_{j=l+1}^L \mW^j \right)^\T 
\ve_\A
\left( \prod_{j=L_f+1}^{l-1} \mW^j \prod_{i=1}^{L_f} \mW^i_\A \right)^\T 
  \nonumber \\
&+ \left( \prod_{j=l+1}^L \mW^j \right)^\T 
\ve_\B
\left( \prod_{j=L_f+1}^{l-1} \mW^j \prod_{i=1}^{L_f} \mW^i_\B \right)^\T   ,
\label{eq:dW}
\end{align}
where the time constant $\tau$ is the inverse of learning rate and $\ve_\A$ and $\ve_\B$ represent the correlation between the output error, $y-\hat y$, and the inputs, $\vx_\A,\vx_\B$,
\begin{subequations}
\begin{align}
\ve_\A &= \mSigma_{y\vx_\A} -  \mW^{\tot}_\A \mSigma_\A -  \mW^{\tot}_\B \mSigma_{\B \A}  , \\
\ve_\B &= \mSigma_{y\vx_\B} - \mW^{\tot}_\A \mSigma_{\A \B} - \mW^{\tot}_\B \mSigma_\B .
\end{align}
\end{subequations}
The network is initialized with small random weights.

\section{Two-Layer Multimodal Linear Networks}
We first study two-layer multimodal linear networks, for which $L=2$. There are two possible fusion schemes for two-layer networks according to our setup: early fusion, $L_f=1$, as in \cref{fig:schematic-early-2L} and late fusion, $L_f=2$, as in \cref{fig:schematic-late-2L}. 

We explain that unimodal bias is conspicuous in late fusion linear networks but not in early fusion linear networks by analyzing their loss landscape in \cref{sec:landscape}. We then focus on late fusion networks: \cref{sec:duration_2L} calculates duration of the unimodal phase; \cref{sec:misattribution} specifies mis-attribution in the unimodal phase; \cref{sec:superficial} reveals the superficial modality preference; \cref{sec:overparam} demonstrates that a long unimodal phase harms generalization in the overparametrized regime.

\subsection{Loss Landscape}  \label{sec:landscape}

\begin{figure*}[ht]
\centering
\subfloat[Network \label{fig:schematic-early-2L}]
{\centering
\includegraphics[width=0.18\linewidth]{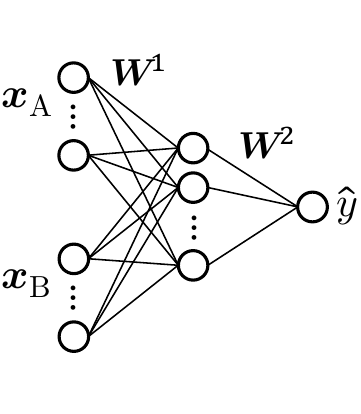}}
\hspace{1ex}
\subfloat[Loss and weights \label{fig:lin_early_rho0_2L}]
{\centering
\includegraphics[width=0.27\linewidth]{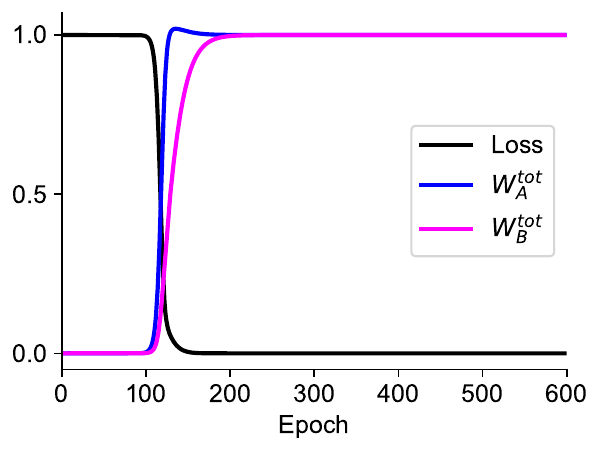}}
\hspace{1ex}
\subfloat[Phase portrait \label{fig:early_phase_portrait}]
{\centering
\includegraphics[width=0.28\linewidth]{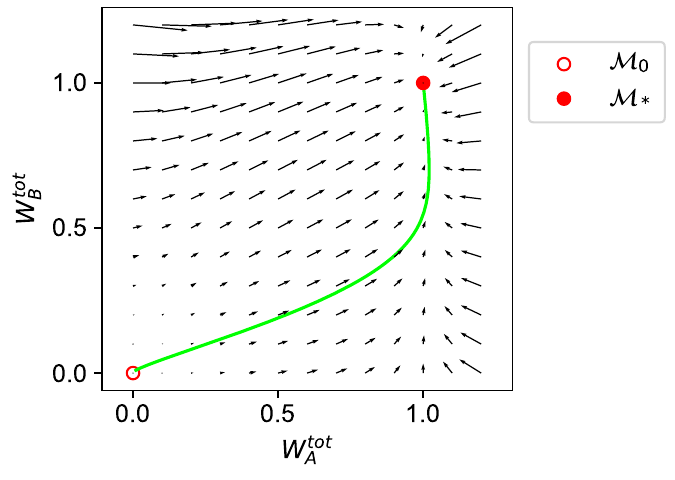}}
\\
\subfloat[Network \label{fig:schematic-late-2L}]
{\centering
\includegraphics[width=0.18\linewidth]{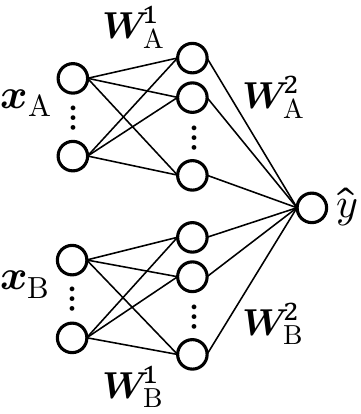}}
\hspace{1ex}
\subfloat[Loss and weights \label{fig:lin_late_rho0_2L}]
{\centering
\includegraphics[width=0.27\linewidth]{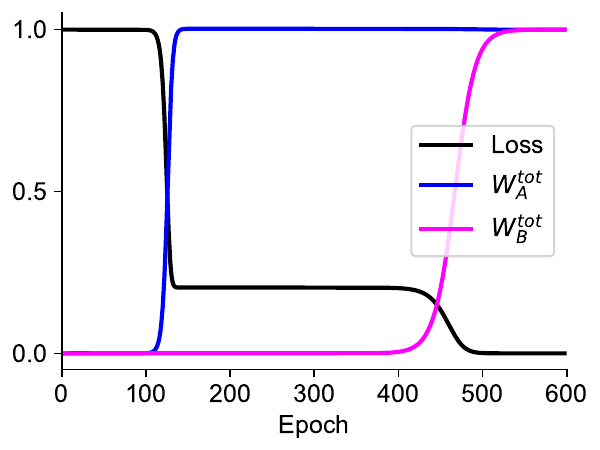}}
\hspace{1ex}
\subfloat[Phase portrait \label{fig:late_phase_portrait}]
{\centering
\includegraphics[width=0.28\linewidth]{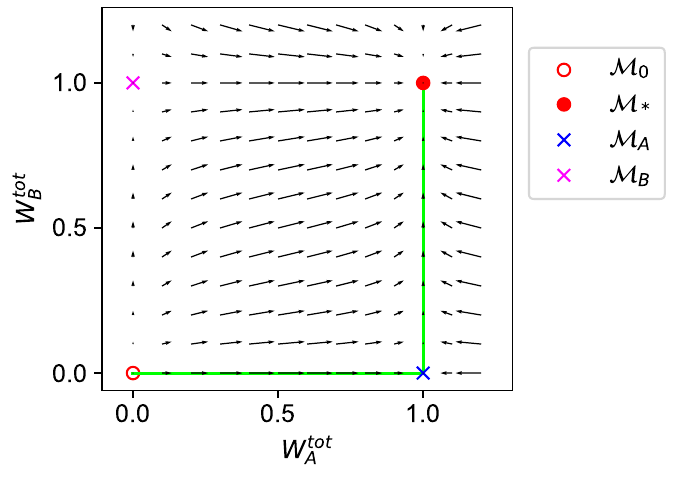}}
\caption{Effect of fusion point on learning dynamics and loss landscape. Top row: Early fusion. Bottom row: Late fusion. Both networks are trained with the same dataset. (a,d) Network schematic. (b,e) Training trajectories. (c,f) Phase portrait. Late fusion introduces two manifolds of saddles (blue and magenta crosses) into the loss landscape, causing learning trajectories to plateau near a unimodal solution. Experimental details are provided in \cref{supp:implementation}.
}
\label{fig:dynamics_2L}
\end{figure*}

As shown in \cref{fig:lin_early_rho0_2L,fig:lin_late_rho0_2L}, early fusion networks learn from both modalities almost simultaneously while late fusion networks learn two modalities at two separate times with a conspicuous unimodal phase in between. For both networks, the loss trajectories exhibit quasi-stage-like behaviors. As studied by \citet{saxe13exact,saxe19semantic,jacot21saddle,pesme23saddle}, linear networks trained from small initialization learn slowly for most of the time and move rapidly from one fixed point or saddle to the next with a sigmoidal transition stage. 
We show that early fusion networks have two manifolds of fixed points, corresponding to their one transition stage. In contrast, late fusion networks have two manifolds of fixed points and two manifolds of saddles, accounting for their two transition stages.

\textbf{Early Fusion}.
There are two manifolds of fixed points in early fusion networks (\cref{supp:early-fixedpoint}): one is an unstable fixed point at zero $\gM_0$ and the other is a manifold of stable fixed points at the global pseudo-inverse solution $\gM_*$,
\begin{subequations}
\begin{align}
\gM_0 &= \{ \mW | \mW=\vzero \}  ; \label{eq:fixedpoint0} \\
\gM_* &= \left\{ \mW | \mW^{\tot} = \mSigma_{y\vx} \mSigma^{-1} \right\}  \label{eq:fixedpointsol} .
\end{align} \label{eq:fixedpoint}%
\end{subequations}
The network starts from small initialization, which is close to the unstable fixed point $\gM_0$. When learning progresses, the network escapes from the unstable fixed point $\gM_0$ and converges to the global pseudo-inverse solution at $\gM_*$ with one brief transition. We visualize the fixed points and the learning trajectory in the phase portrait in \cref{fig:early_phase_portrait}.
Since there can only be one brief transition stage, all modalities are learned almost simultaneously in early fusion networks. 

\textbf{Late Fusion}.
Late fusion linear networks have the same two manifolds of fixed points $\gM_0,\gM_*$ as early fusion networks.
In addition, late fusion linear networks have two manifolds of saddles $\gM_\A, \gM_\B$ (\cref{supp:late-fixedpoint}), corresponding to learning one modality but not the other,
\begin{subequations}
\begin{align}
\gM_\A &= \left\{ \mW | \mW^{\tot}_\A=\mSigma_{y \vx_\A} {\mSigma_\A}^{-1}, \mW^{\tot}_\B = \vzero \right\}  ; \label{eq:saddleA} \\
\gM_\B &= \left\{ \mW | \mW^{\tot}_\A=\vzero, \mW^{\tot}_\B = \mSigma_{y \vx_\B} {\mSigma_\B}^{-1} \right\}  \label{eq:saddleB} .
\end{align}  \label{eq:saddles}%
\end{subequations}
The late fusion linear network therefore undergoes two transition stages because the network first arrives and lingers near a saddle in $\gM_\A$ or $\gM_\B$ and subsequently converges to the global pseudo-inverse solution in $\gM_*$. Visiting the saddle gives rise to the plateau in the loss that separates the time when the two modalities are learned. During the plateau, the network is unimodal. We visualize the fixed points and the learning trajectory in the phase portrait in \cref{fig:late_phase_portrait}.

\begin{figure*}
\centering
\subfloat[$\rho=0.5$ loss and weights \label{fig:overshoot}]
{\centering
\includegraphics[width=0.25\linewidth]{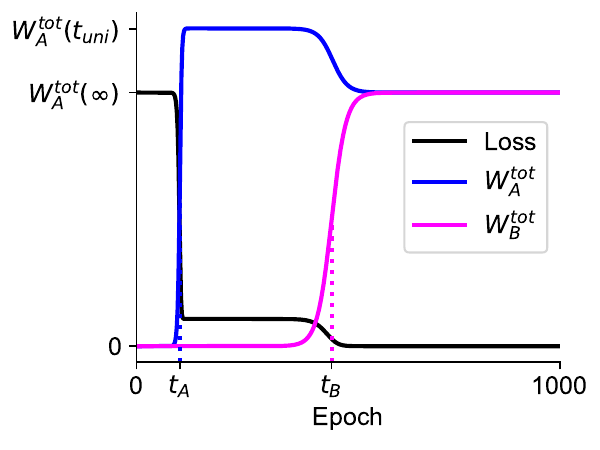}}
\subfloat[$\rho=-0.5$ loss and weights \label{fig:undershoot}]
{\centering
\includegraphics[width=0.25\linewidth]{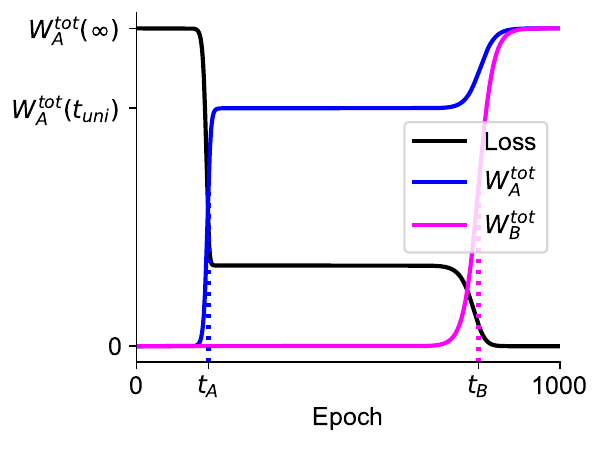}}
\subfloat[Time ratio \label{fig:timesweep_2L}]
{\centering
\includegraphics[width=0.25\linewidth]{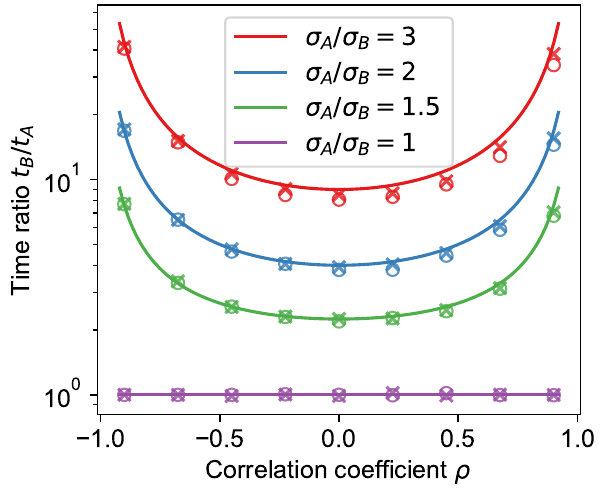}}
\subfloat[Mis-attribution \label{fig:amountsweep_2L}]
{\centering
\includegraphics[width=0.25\linewidth]{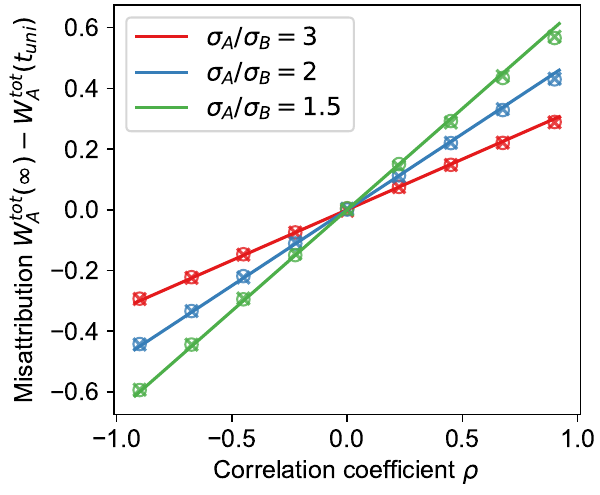}}
\caption{Duration of unimodal phase and amount of mis-attribution in two-layer late fusion linear networks. We consider scalar inputs $\vx_\A,\vx_\B \in \sR$ with input covariance matrix parameterized as $\mSigma = \left[\sigma_\A^2, \rho \sigma_\A \sigma_\B; \rho \sigma_\B \sigma_\A, \sigma_\B^2 \right]$. The target output is generated as $y=\vx_\A+\vx_\B$.
(a) Loss and total weight trajectories in two-layer late fusion networks when modalities are positively correlated. (b) Same as panel a but for negative correlations.
(c) Time ratio $t_\B/t_\A$ as in \cref{eq:timeratio-2L}. 
(d) Amount of mis-attribution. In panel c and d, lines are theoretical predictions; circles are simulations of two-layer late fusion linear networks; crosses are simulations of two-layer late fusion ReLU networks. Experimental details are provided in \cref{supp:implementation}.}
\label{fig:sweep_2L}
\end{figure*}

\subsection{Duration of the Unimodal Phase} \label{sec:duration_2L}
We now quantify the duration of the unimodal phase. For definiteness, we designate the modality learned first to be modality $\A$. We define $t_\A$ to be the time when the total weight of modality A reaches half of its associated plateau and similarly for $t_\B$, as illustrated in \cref{fig:overshoot,fig:undershoot}.
Because of the small random initialization, we can assume that the Frobenius norms of $\mW_\A^1$ and $\mW_\B^1$ are approximately equal at initialization, denoted $u_0$. 
We compute the times $t_\A$ and $t_\B$ in \cref{supp:time-2L} in leading order of the initialization $u_0$ and obtain
\begin{subequations}
\begin{align} 
t_\A &= \tau \| \mSigma_{y \vx_\A} \|^{-1} \ln \frac1{u_0} 
, \\
t_\B &= t_\A + \tau \frac{1 - \| \mSigma_{y \vx_\A} \|^{-1} \| \mSigma_{y \vx_\B}\|} {\lnorm \mSigma_{y \vx_\B} - \mSigma_{y \vx_\A} {\mSigma_\A}^{-1} \mSigma_{\A \B} \rnorm} \ln \frac1{u_0}  .
\end{align} 
\label{eq:tAtB-2L}%
\end{subequations}
To compare the unimodal phase duration across different settings, we focus on the time ratio\footnote{We use $\| \cdot \|$ to notate the L2 norm of a vector or the Frobenius norm of a matrix in this paper.},
\begin{align} \label{eq:timeratio-2L}
\frac{t_\B}{t_\A} = 1 + \frac{\| \mSigma_{y \vx_\A} \| - \| \mSigma_{y \vx_\B} \|} {\lnorm \mSigma_{y \vx_\B} - \mSigma_{y \vx_\A} {\mSigma_\A}^{-1} \mSigma_{\A \B} \rnorm} .
\end{align}
We note that the time ratio reduces to $\| \mSigma_{y \vx_\A} \| / \| \mSigma_{y \vx_\B} \|$ if the cross correlation is zero, i.e., $\mSigma_{\A \B}=\vzero$. This accords with the intuition that $\| \mSigma_{y \vx_\A} \|$ governs the speed at which modality A is learned and $\| \mSigma_{y \vx_\B} \|$ governs the speed at which modality B is learned. 
When the cross correlation is nonzero, having learned modality A affects the speed at which modality B is learned during the unimodal phase. Specifically, the speed of modality B is reduced by $\mSigma_{y \vx_\A} {\mSigma_\A}^{-1} \mSigma_{\A \B}$; see \cref{supp:time-2L}. 

We validate \cref{eq:timeratio-2L} with numerical simulations in \cref{fig:timesweep_2L}. With theoretical and experimental evidence, we conclude that stronger correlations between input modalities and a greater disparity in input-output correlations for each modality make the time ratio larger, indicating a longer unimodal phase. In the extreme case of maximum correlation, $\vx_\A$ and $\vx_\B$ are co-linear, and so one of them is redundant. In this case, the denominator $\mSigma_{y \vx_\B} - \mSigma_{y \vx_\A} {\mSigma_\A}^{-1} \mSigma_{\A \B}$ is 0, and the ratio $t_\B/t_\A$ is $\infty$. Here later becomes never --- the network learns to fit the output only with modality A and modality B will never be learned, as shown in \cref{fig:collinear}. 

\subsection{Mis-attribution in the Unimodal Phase}  \label{sec:misattribution}
We take a closer look at the unimodal mode by highlighting a phenomenon we call mis-attribution. 
During the unimodal phase, $\mW^{\tot}_\A$ fits the output as much as it can and the network mis-attributes some of the output contributed by modality B to modality A by exploiting their correlations.
When modalities are correlated, the local pseudo-inverse solution differs from the global pseudo-inverse solution, i.e., $\left[ \mSigma_{y \vx_\A} {\mSigma_\A}^{-1}, \mSigma_{y \vx_\B} {\mSigma_\B}^{-1} \right] \neq \mSigma_{y \vx} {\mSigma}^{-1}$. 
Specifically, the weights of modality $\A$ overshoot if modalities have a positive correlation as in \cref{fig:overshoot} and undershoot if negative as in \cref{fig:undershoot}. This mis-attribution is then corrected when modality $\B$ catches up at time $t_\B$, and the network eventually converges to the global pseudo-inverse solution. In \cref{fig:amountsweep_2L} we demonstrate, using scalar input for clarity, that mis-attribution is more severe when modalities have stronger correlation.

When modalities are uncorrelated, late fusion networks do not mis-attribute during the unimodal phase because the local pseudo-inverse solutions are the same as the global pseudo-inverse solution. Weights for modality A converge to the global solution at time $t_\A$ and do not change thereafter, as in \cref{fig:lin_late_rho0_2L}. In this case, the late fusion network behaves the same as two separately trained unimodal networks.

\begin{figure}
\centering
\subfloat[Superficial preference \label{fig:superficial-ex}]
{\centering
\includegraphics[width=0.5\linewidth]{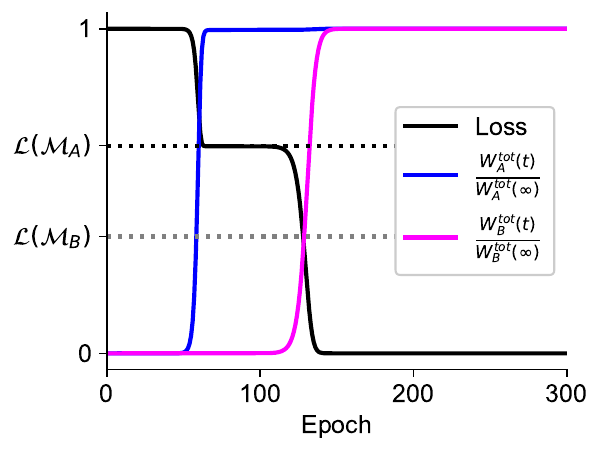}}
\subfloat[Non-superficial preference \label{fig:superficial-counterex}]
{\centering
\includegraphics[width=0.5\linewidth]{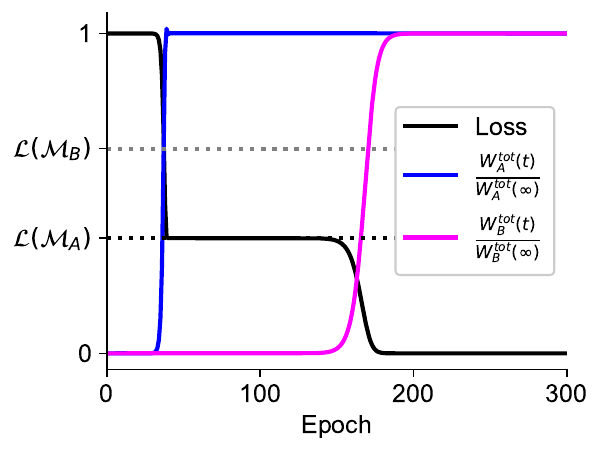}}
\\
\subfloat[Superficial region \label{fig:superficial-region}]
{\centering
\includegraphics[width=0.5\linewidth]{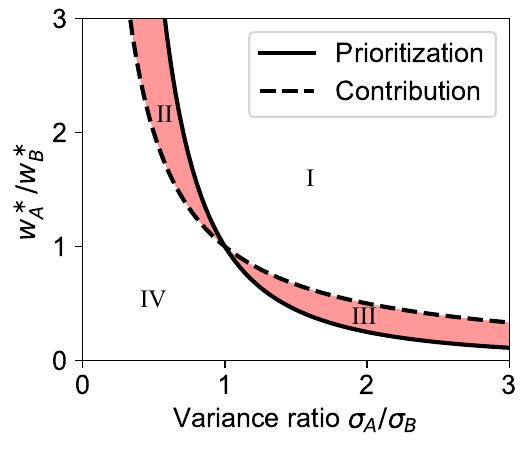}}
\caption{Demonstration of superficial modality preference. A two-layer late fusion linear network is trained with two different dataset. (a,b) In both examples, modality A is learned first. The dotted black line marks the loss when the network visits $\gM_\A$. The dotted gray line marks the loss if the network had instead visited $\gM_\B$. (a) The prioritized modality is not the modality that contributes more to the output. (b) The prioritized modality is the modality that contributes more to the output. (c) Boundaries of which modality is prioritized and which modality contributes more to the output in terms of dataset statistics. In region I and III, modality A is learned first. In region I and II, modality A contributes more to the output. Thus in region II and III (shaded red), prioritization and contribution disagree, resulting in superficial modality preference.
Experimental details are provided in \cref{supp:implementation}.
}
\label{fig:superficial}
\end{figure}

\subsection{Superficial Modality Preference}  \label{sec:superficial}
We now look into which modality is learned first. Late fusion linear networks have what we call ``superficial modality preference''. They prioritize the modality that is faster to learn, which is not necessarily the modality that yields the larger decrease in loss. 

From the time ratio expression given in \cref{eq:timeratio-2L} and details in \cref{supp:prioritization}, we see that which modality is learned first depends solely on the relative size of $\| \mSigma_{y \vx_\A} \|$ and $\| \mSigma_{y \vx_\B} \|$.
Thus late fusion networks first learn the modality that has a higher correlation with the output, even though it may not be the modality that makes a larger contribution to the output. Under the following two conditions on dataset statistics,
\begin{subequations}  \label{eq:superficial}
\begin{align} 
\| \mSigma_{y \vx_\A} \| &> \| \mSigma_{y \vx_\B} \|  , \\
\mSigma_{y \vx_\A} {\mSigma_\A}^{-1} \mSigma_{y \vx_\A}^\T &< \mSigma_{y \vx_\B} {\mSigma_\B}^{-1} \mSigma_{y \vx_\B}^\T  ,
\end{align}
\end{subequations}
modality A is faster to learn but modality B contributes more to the output.
We present an example where these conditions hold in \cref{fig:superficial-ex} and where they do not in \cref{fig:superficial-counterex}.

If $\vx_\A,\vx_\B$ are scalars and uncorrelated, the two inequality conditions in \cref{eq:superficial} reduce to 
\begin{align}  \label{eq:superficial-scalar}
\frac{\sigma_\B^2}{\sigma_\A^2} < \frac{w_\A^\ast}{w_\B^\ast} < \frac{\sigma_\B}{\sigma_\A}  ,
\end{align}
where $\sigma_\A,\sigma_\B$ are variances of $\vx_\A,\vx_\B$ and we assume the target output is generated as $y=w_\A^\ast \vx_\A + w_\B^\ast \vx_\B$. We plot the two conditions in \cref{fig:superficial-region}. Region III satisfies the conditions we give in \cref{eq:superficial-scalar}. Region II corresponds to \cref{eq:superficial-scalar} with flipped inequality signs, meaning the other superficial modality preference case where modality B is prioritized but modality A contributes more to the output. Hence, region II and III (shaded red) cover the dataset statistics where prioritization and contribution disagree and late fusion linear networks would prioritize learning the modality that contributes less to the output. 

We note that the networks eventually converge to zero loss, regardless of which modality is learned first and how long the unimodal phase is. But they affect the generalization error in the overparameterized regime, which we discuss in the following section.

\subsection{Underparameterization and Overparameterization}  \label{sec:overparam}

\begin{figure}[h]
\centering
\subfloat[Underparam early fusion \label{fig:underparam-early-Eg}]
{\centering
\includegraphics[width=0.5\linewidth]{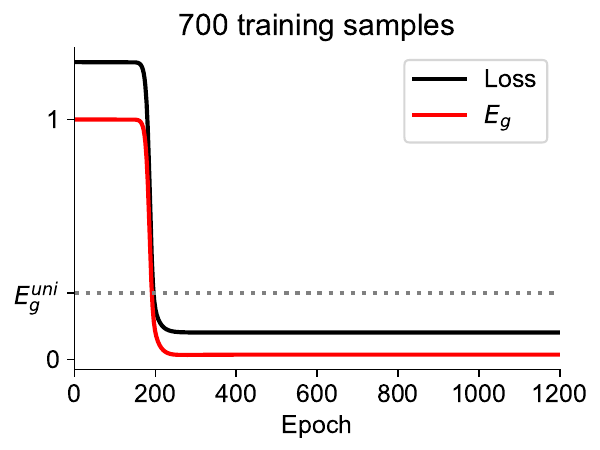}}
\subfloat[Underparam late fusion \label{fig:underparam-late-Eg}]
{\centering
\includegraphics[width=0.5\linewidth]{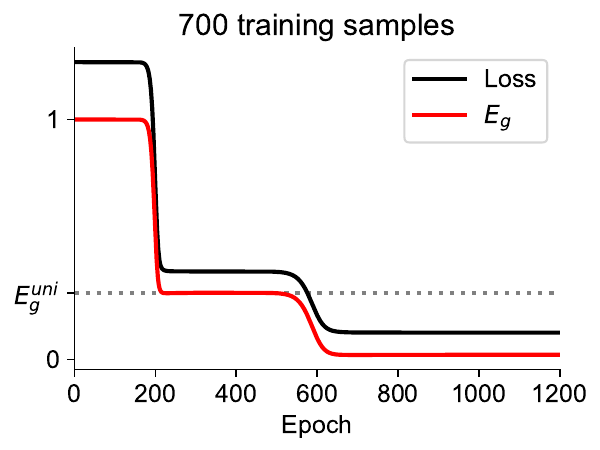}}
\\
\subfloat[Overparam early fusion \label{fig:overparam-early-Eg}]
{\centering
\includegraphics[width=0.5\linewidth]{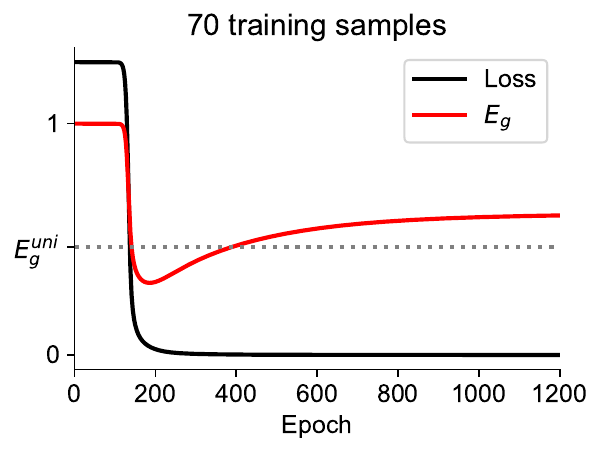}}
\subfloat[Overparam late fusion \label{fig:overparam-late-Eg}]
{\centering
\includegraphics[width=0.5\linewidth]{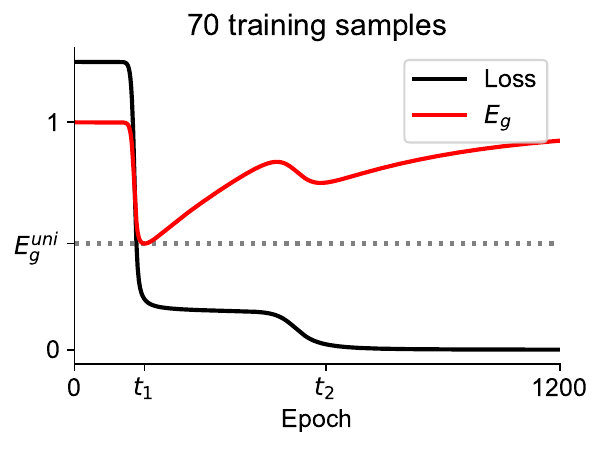}}
\caption{Overparameterized and underparameterized two-layer early and late fusion linear networks. Inputs are 50-dimensional, i.e., $\vx_\A,\vx_\B \in \sR^{50}$. 
(a) Loss and generalization error trajectories of a two-layer early fusion linear network trained with 700 examples. (b) Same as panel a but with late fusion. (c) Loss and generalization error trajectories of a two-layer early fusion linear network trained with 70 examples. (d) Same as panel c but with late fusion. The dotted gray line marks the lowest generalization error that a unimodal network could achieve with the same dataset.
Experimental details are given in \cref{supp:implementation}.}
\label{fig:Eg}
\end{figure}

In the underparameterized regime, training loss closely tracks the corresponding generalization error as shown in \cref{fig:underparam-early-Eg,fig:underparam-late-Eg} because the training data is sufficient to accurately estimate the true data distribution. Analysis on the training loss applies to the generalization error as well. Both early and late fusion networks achieve a lower generalization error at convergence than that of a unimodal network (dotted gray line).

In the overparameterized regime, the number of samples is insufficient compared to the number of effective parameters, which is the input dimension for linear networks \citep{advani20highd}. As shown in \cref{fig:overparam-early-Eg}, the overparameterized early fusion linear network learns both modalities during one transition stage. The generalization error decreases during the transition stage and increases afterwards, as predicted by theory \citep{lecun91eigen,krogh92gen,advani20highd}. If early stopping is adopted, we obtain a model that has learned from both modalities and does not substantially overfit. This model achieves a lower generalization error than its unimodal counterpart.

As shown in \cref{fig:overparam-late-Eg}, the overparameterized late fusion linear network learns the faster-to-learn modality first and overfits this modality during the unimodal phase when the training loss plateaus but the generalization error increases. In this case, there is a dilemma between overfitting the first modality and underfitting the second. 
If optimal early stopping on the generalization error is adopted, training terminates at time $t_1$. The model at time $t_1$ has a strong and permanent unimodal bias because the first modality has just been learned while the second modality has not.
If training terminates at time $t_2$, after both modalities are learned, the network has overfit the first modality. This can result in a generalization error worse than that in early stopping or what could be achieved by its unimodal counterpart.
Thus overfitting is a mechanism that can convert the transient unimodal phase to a generalization deficit and permanent unimodal bias.

\section{Deep Multimodal Linear Networks}
We now consider the more general case of multimodal deep linear networks and examine how the fusion layer depth $L_f$ affects the extent of unimodal bias.

\begin{figure*}
\centering
\subfloat[Loss \label{fig:dynamics-4L}]
{\centering
\includegraphics[width=0.23\linewidth]{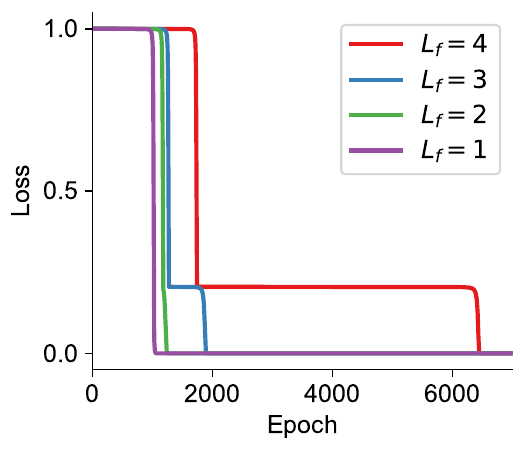}}
\subfloat[Correlation sweep \label{fig:rho-sweep-4L}]
{\centering
\includegraphics[width=0.26\linewidth]{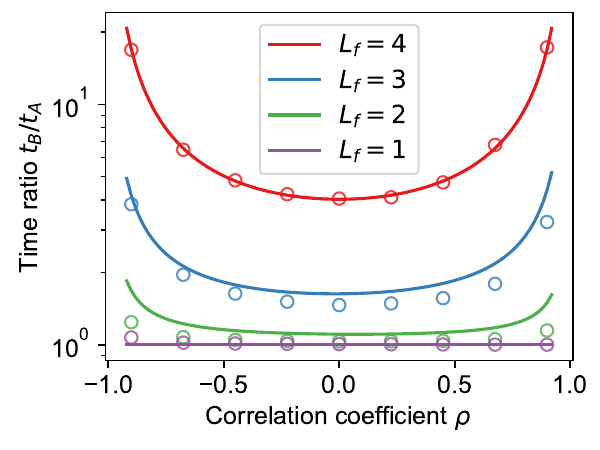}}
\subfloat[Variance ratio sweep \label{fig:ratio-sweep-4L}]
{\centering
\includegraphics[width=0.26\linewidth]{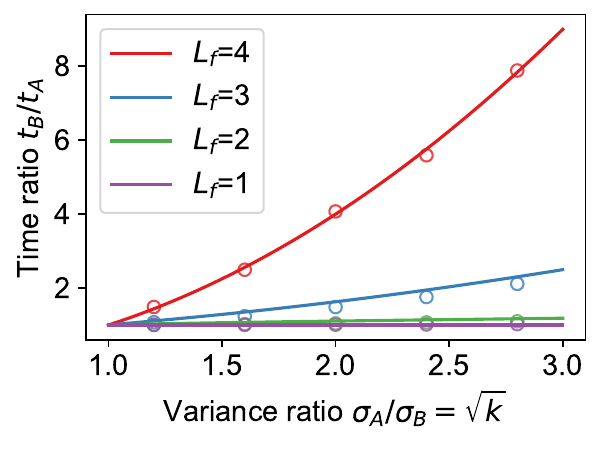}}
\subfloat[Initialization sweep \label{fig:init-sweep-4L}]
{\centering
\includegraphics[width=0.26\linewidth]{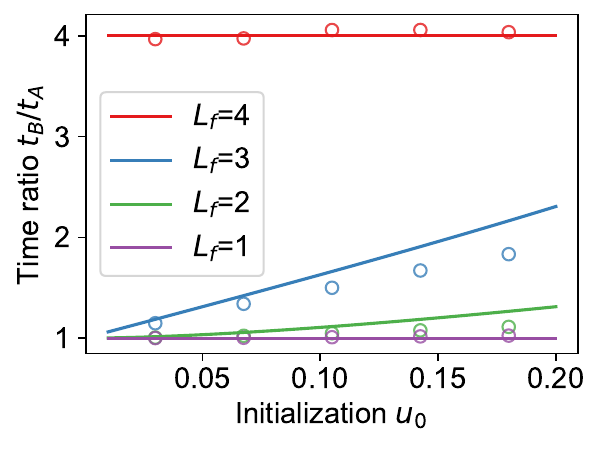}}
\caption{Duration of unimodal phase in multimodal deep linear networks. Four-layer linear networks with fusion layer at $L_f=1,2,3,4$ are trained to fit $y=\vx_\A + \vx_\B$. The input covariance matrix is parameterized as $\mSigma = \left[\sigma_\A^2, \rho \sigma_\A \sigma_\B; \rho \sigma_\B \sigma_\A, \sigma_\B^2 \right]$.
(a) An example of loss trajectories when input covariance matrix $\mSigma=\text{diag}(4,1)$.
(b) Correlation coefficient sweep with $\sigma_\A/\sigma_\B=2$ and initialization scale $u_0=0.1$. (c) Variance ratio sweep with $\rho=0, u_0=0.1$. Note that $\sigma_\A/\sigma_\B=\sqrt k$ when $\rho=0$. (d) Initialization scale sweep with $\sigma_\A/\sigma_\B=2, \rho=0$. In panel b to d, lines are theory; circles are simulations of four-layer linear networks with different fusion layer depth. Experimental details are provided in \cref{supp:implementation}.}
\label{fig:depth4_sweep}
\end{figure*}

\subsection{Loss Landscape}
\textbf{Early Fusion}.
Similar to two-layer early fusion networks, deep early fusion networks learn from both modalities almost simultaneously, as shown in \cref{fig:dynamics-4L} (purple curve).
\citet{saxe13exact,saxe19semantic,advani20highd} have shown that for deep linear networks, depth slows down learning but does not qualitatively change the dynamics compared to two-layer linear networks. 
The weights associated with all input modalities escape from the initial zero fixed point $\gM_0$ and converge to the pseudo-inverse solution fixed point in $\gM_*$ in one transition stage. 

\textbf{Intermediate and Late Fusion}.
Deep intermediate and late fusion linear networks learn the two modalities with two separate transitions, as shown in \cref{fig:dynamics-4L}; this is similar to what happens in two-layer late fusion networks. Due to the common terms in \cref{eq:dWAdWB,eq:dW} that govern the weight dynamics, the two manifolds of fixed points, $\gM_0, \gM_*$, and the two manifolds of saddles, $\gM_\A, \gM_\B$, exist for any $2\leq L_f\leq L, L\geq2$, comprising intermediate and late fusion linear networks of any configuration.

In what follows, we stick to the convention that modality A is learned first. Deep intermediate and late fusion networks start from small initialization, which is close to the zero fixed point $\gM_0$. When modality A is learned in the first transition stage at time $t_\A$, the network visits the saddle in $\gM_\A$. After a unimodal phase, the network goes through the second transition at time $t_\B$ to reach the global pseudo-inverse solution fixed point in $\gM_*$. Because the network is in the same manifold $\gM_\A$ during the unimodal phase, our results in Section \ref{sec:misattribution} on the mis-attribution in the unimodal phase and Section \ref{sec:superficial} on superficial modality preference in two-layer late fusion networks directly carry over to deep intermediate and late fusion linear networks. 

As shown in \cref{fig:depth4_sweep}, the loss trajectories of networks with different $L_f\geq 2$ traverse the same plateau but they stay in the plateau for different durations. We thus quantify how the total depth $L$ and fusion layer depth $L_f$ affect the duration of the unimodal phase in \cref{sec:duration_deep}.

\subsection{Duration of the Unimodal Phase}  \label{sec:duration_deep}
We now calculate the duration of the unimodal phase in deep intermediate and late fusion linear networks, incorporating the new parameters $L$ and $L_f$. The input-output correlation ratio is denoted $k=\| \mSigma_{y\vx_\B} \| / \| \mSigma_{y\vx_\A} \| \in (0,1)$. We derive the time ratio through leading order approximation of the initialization scale in \cref{supp:time-deep}. For $2<L_f\leq L$, the time ratio is
\begin{align}  \label{eq:deep-late-timeratio}
\begin{split}
\frac{t_\B}{t_\A} = 1 &+ \frac{\| \mSigma_{y \vx_\A} \|-\| \mSigma_{y \vx_\B} \|}{\lnorm \mSigma_{y \vx_\B} - \mSigma_{y \vx_\A} {\mSigma_\A}^{-1} \mSigma_{\A \B} \rnorm}   \\
&\times \frac{u_0^{L-L_f}} {(L_f-2) \| \mSigma_{y \vx_\A} {\mSigma_\A}^{-1} \|^{1-\frac{L_f}L} } I(L, L_f)^{-1}  ,
\end{split}
\end{align}
where 
\begin{align}  \label{eq:integral}
&I(L, L_f) = \int_{1} ^ \infty \frac{\left[ 1 + \left( k + \left(1-k \right) x^{L_f-2} \right)^{\frac2{2-L_f}} \right]^{\frac{L_f-L}2}}
{x^{L-1}}  dx
\, .
\end{align}
For $L_f=2$, the expression is slightly different but qualitatively similar; see \cref{supp:time-deep-2Lf}.
As shown in \cref{fig:depth4_sweep}, the theoretical prediction captures the trend that a deeper fusion layer $L_f$, a larger input-output correlation ratio $k$, and stronger correlations $\mSigma_{\A \B}$ between input modalities all prolong the duration of the unimodal phase. The qualitative influence of dataset statistics on the time ratio in deep intermediate and late fusion networks is consistent with what we have seen in two-layer late fusion networks.

We now look into the influence of the fusion layer depth.
By setting $L_f=L$ in \cref{eq:deep-late-timeratio}, we find that the time ratio in deep late fusion networks reduces to the same expression as the two-layer late fusion case in \cref{eq:timeratio-2L}, which only involves dataset statistics but not the depth of the network or the initialization, since depth slows down the learning of both modalities by the same factor. 
In intermediate fusion linear networks, the time ratio is smaller than in late fusion networks, with a smaller ratio for a shallower fusion layer. In intermediate fusion linear networks, learning one modality changes the weights in its associated pre-fusion layers and the shared post-fusion layers. At time $t_\A$, the pre-fusion layer weights of modality A and the shared post-fusion layer weights have escaped from the zero fixed point and grown in scale while the pre-fusion layer weights of modality B have not. During the unimodal phase, the shared post-fusion layer weights and the correlation between modality B and the output together drive the pre-fusion layer weights of modality B to escape from the zero fixed point. Thus having more shared post-fusion layers makes learning one modality more helpful for learning the other, shortening the unimodal phase. In essence, an early fusion point allows the weaker modality to benefit from the stronger modality's learning in the post-fusion layers.

We also note that the initialization scale affects the time ratio in intermediate fusion networks, as demonstrated in \cref{fig:init-sweep-4L}. 
Even amongst cases that all fall into the rich feature learning regime, the initialization scale has an effect on the time ratio, with a larger time ratio for a larger initialization scale.
In \cref{fig:init-sweep-4L}, the simulations (circles) slightly deviate from theoretical predictions (lines) because our theoretical prediction is derived with small initialization and is thus less accurate for larger initialization. Nonetheless, the monotonic trend is well captured.

In summary, a deeper fusion layer, a larger input-output correlation ratio, stronger correlations between input modalities, and sometimes a smaller initialization scale all prolong the unimodal phase in the joint training of multimodal deep linear networks with small initialization.

\section{Discussion}
To understand unimodal bias in multimodal learning, we have studied multimodal deep linear networks. In these networks, the unimodal bias already manifests itself in many ways consistent with those seen in complex networks.
We now discuss how our results can apply to or break in nonlinear networks and tasks, in the hope of inspiring future work.

\subsection{Nonlinear Multimodal Networks}
\textbf{Linear Task}. 
We find that our results, derived for linear networks, carry over to two-layer ReLU networks when the target task is linear. This aligns with the intuitions from a line of studies on the implicit bias of two-layer ReLU networks \citep{sarussi21relu,phuong21relu,timor23relu,min24relu}. 
We simulate two-layer early and late fusion ReLU networks to learn a linear target map. As shown in \cref{fig:ReLU_2L}, the loss and weight trajectories are qualitatively the same as their linear counterparts in \cref{fig:dynamics_2L}, except that learning is about two times slower and the converged total weights are two times larger.
We conduct simulations on the duration of the unimodal phase and the amount of mis-attribution in two-layer late fusion ReLU networks with the rest of the setting unchanged (target map is linear). The time ratio and mis-attribution in ReLU networks closely follow the theoretical predictions derived for linear networks as shown in \cref{fig:timesweep_2L,fig:amountsweep_2L} (crosses).

\begin{figure*}
\centering
\subfloat[$L_f=1$]
{\centering
\includegraphics[width=0.18\linewidth]{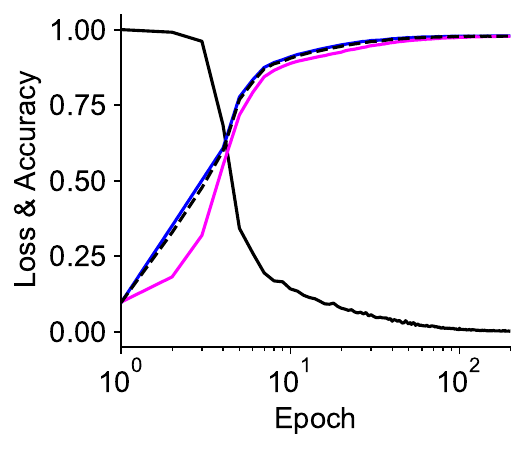}}
\subfloat[$L_f=2$]
{\centering
\includegraphics[width=0.18\linewidth]{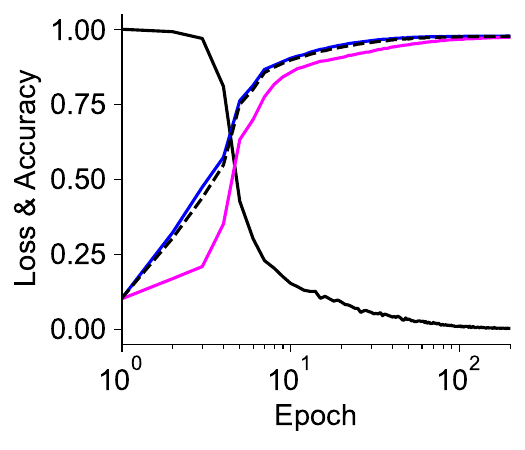}}
\subfloat[$L_f=3$]
{\centering
\includegraphics[width=0.18\linewidth]{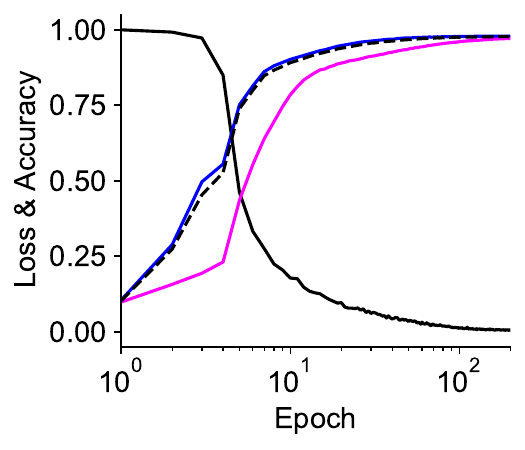}}
\subfloat[$L_f=4$]
{\centering
\includegraphics[width=0.18\linewidth]{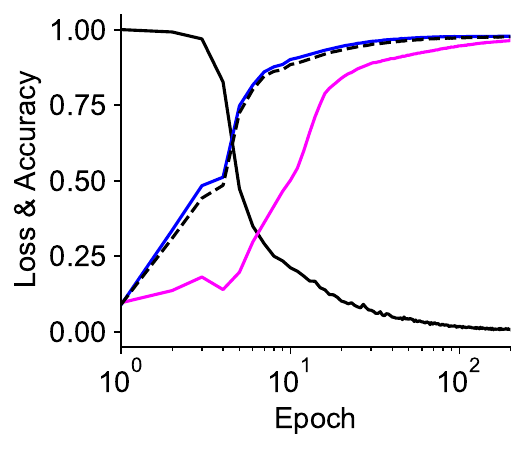}}
\subfloat[$L_f=5$]
{\centering
\includegraphics[width=0.267\linewidth]{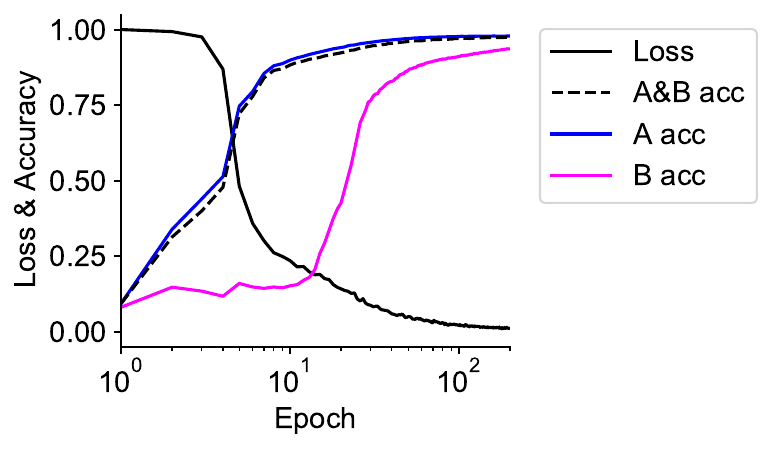}}
\caption{Loss and accuracy trajectories of multimodal fully-connected ReLU networks. Five-layer ReLU newtorks with different fusion layer depths are trained on the noisy MNIST dataset. Solid black curves plot the training loss. Dotted black curves plot the multimodal classification accuracy, where both branches are presented with an uncorrupted testset image. Blue curves plot the unimodal classification accuracy of modality A, where branch A is presented with an uncorrupted testset image and branch B with a blank image. Similarly, for modality B (purple curves). The trajectories are averaged over five random seeds. Experimental details are provided in \cref{supp:mnist}.}
\label{fig:fcn-mnist}
\end{figure*}

\textbf{Realistic Task}.
We validate our results in multimodal deep ReLU networks trained on a noisy MNIST \citep{lecun98MNIST} task. In the noisy MNIST classification task shown in \cref{fig:mnist-sample}, the multimodal network receives two MNIST written digit images as input, one of which may be corrupted by Gaussian noise. This is a common scenario in multimodal learning where the dominating modality varies per sample. 
We train multimodal ReLU networks with $L=5$ and $L_f=1,2,\cdots,5$ on this task. The multimodal ReLU networks learn the modality that is corrupted less often faster. We record the ratio between the time when the unimodal accuracy of the first learned modality reaches $50\%$ and the time of the second modality. \cref{table} records the mean and standard deviation of the time ratio across five random seeds. As shown in \cref{fig:fcn-mnist,table}, the deeper the layer at which fusion occurs, the slower the network learns from the second modality. This is consistent with the qualitative conclusion we draw from multimodal linear networks. Similar results are observed in multimodal convolutional networks, as shown in \cref{fig:cnn-mnist,table}.
\begin{table}[h]
\caption{Time ratio in noisy MNIST digit classification.}
\vskip 0.15in
\begin{center}
\begin{small}
\begin{tabular}{lccccc}
\toprule
Network & MLP & CNN  \\
\midrule
$L_f=1$ & 1.30$\pm$0.24 & 1.04$\pm$0.03  \\
$L_f=2$ & 1.53$\pm$0.40 & 1.10$\pm$0.05  \\
$L_f=3$ & 1.90$\pm$0.39 & 1.08$\pm$0.03  \\
$L_f=4$ & 3.38$\pm$1.13 & 1.32$\pm$0.16  \\
$L_f=5$ & 6.38$\pm$1.12 & 2.53$\pm$0.45  \\
\bottomrule
\end{tabular}
\end{small}
\end{center}
\vskip -0.1in
\label{table}
\end{table}

\textbf{Heterogeneous Task}. 
We present a simple heterogeneous task that involves behaviors not observed in linear networks and linear tasks.
Consider learning $y=\vx_\A + \text{XOR}(\vx_\B)$, where $\vx_\A \in \sR, \vx_\B \in \{ [1,1], [1,-1],[-1,1], [-1,-1] \}$ and $\text{XOR}(\vx_\B)$ refers to performing XOR to the two dimensions of $\vx_\B$. We observe that two-layer late fusion ReLU networks always learn this task successfully, forming the four perpendicular XOR features as shown in \cref{fig:xor-late-1var,fig:xor-late-2var,fig:xor-late-3var}. However, two-layer early fusion ReLU networks do not learn consistent XOR features and can even fail to learn this task as shown in \cref{fig:xor-early-1var,fig:xor-early-2var,fig:xor-early-3var}. In the failed cases, the variance of $\vx_\A$ is large so that the network can be stuck at a local minimum where the network only exploits the linear modality. For this heterogeneous task, late fusion networks are advantageous in terms of extracting heterogeneous features from each input modality. We provide videos of the learning dynamics on our website.

\subsection{Practical Insights}
We hypothesize that the practical choice of the fusion layer depth is a trade-off between alleviating unimodal bias and learning unimodal features. A shallower fusion layer helps alleviate unimodal bias because modalities can cooperate reciprocally to learn the synergistic computation. A deeper fusion layer helps unimodal feature learning because the network can operate more independently to learn the unique computation of extracting heterogeneous features from each modality. We hope our work contributes to a better understanding of this trade-off, ultimately leading to more systematic architectural choices and improved multimodal learning algorithms.

\section*{Acknowledgements}
We thank Loek van Rossem and William Dorrell for feedback on a draft of this paper.
We thank the following funding sources: Gatsby Charitable Foundation to YZ, PEL, and AS; Wellcome Trust (110114/Z/15/Z) to PEL; Sainsbury Wellcome Centre Core Grant from Wellcome (219627/Z/19/Z) to AS; a Sir Henry Dale Fellowship from the Wellcome Trust and Royal Society (216386/Z/19/Z) to AS.

\section*{Impact Statement}
This paper presents work whose goal is to advance the field of multimodal deep learning. There are many potential societal consequences of our work, none of which we feel must be specifically highlighted here.

\bibliography{references}
\bibliographystyle{icml2024}

\newpage
\appendix
\onecolumn

\begin{center}
{\LARGE\bfseries Appendix}
\end{center}

\section{Additional Related Work}
\subsection{Technical Challenges in Multimodal Deep Linear Networks}
There are two main challenges of studying multimodal deep linear networks, compared with standard deep linear networks.

The first challenge is that multimodal networks are not fully-connected. 
The gradient descent dynamics of multimodal linear networks derived in \cref{supp:grad} differ from those in fully-connected linear networks. One complication is that the standard balancing property \citep{du18autobalance} does not hold across all layers. In fully-connected linear networks, all layers maintain balance and have equal norm throughout learning \citep{du18autobalance}. However, in multimodal linear networks, post-fusion layers are balanced and pre-fusion layers of the two modalities are respectively balanced, as discussed in \cref{supp:balancing}. Thus, the dynamics of multimodal linear networks are summarized by three variables (post-fusion layers, pre-fusion layers of modality A, and pre-fusion layers of modality B), in contrast to one variable in fully-connected linear networks.

The second challenge is that multimodal datasets generally do not have whitened input.
A considerable portion of theoretical work on linear network learning dynamics \citep{fukumizu98batch,saxe13exact,saxe19semantic,lampinen2018gen,arora19converge,huh20curvature,tarmoun21overparam,cengiz22silent,clem22prior} assumes whitened input covariance matrix, i.e. identity matrix. With the whitened input assumption, the learning dynamics of linear networks with small or balanced initialization can be written as a Riccati equation, which has known closed-form solutions \citep{fukumizu98batch,saxe13exact,tarmoun21overparam,shi22pathways,clem22prior}.
However, we did not use this assumption because it is unrealistic to assume that multimodal data streams are whitened. For the unwhitened input case, existing literature does not provide closed-form solutions and \citet{tarmoun21overparam} stated that they believe the dynamics cannot be solved exactly. Given the unavailability of closed-form solutions, we conduct a fixed point analysis and a calculation of the time ratio to gain insights relevant to unimodal bias. Our fixed point analysis is a standard tool for dynamical systems but has not been done for multimodal deep linear networks before. Our calculation of the time presents novel tools to understand the timing of phase transitions in linear networks.

\subsection{Alleviating Unimodal Bias in Practice}
Unimodal bias is a common issue in multimodal learning. Practitioners have proposed many approaches to alleviate the bias.

Some approaches construct more balanced datasets to alleviate the language bias in Visual Question Answering (VQA). \citet{zhang16yinyang} added training samples with the opposite answer for binary VQA datasets. \citet{goyal17cvpr} built VQA v2 dataset by adding images with different answers into the popular VQA v1 dataset \citep{antol15VQA}. \citet{agrawal18cvpr} presented new splits of the VQA v1 and VQA v2 datasets in which the train and test sets have different prior distributions of answers. The new splits can be used to diagnose the unimodal bias in VQA models.

Some approaches alter the training process of multimodal models to alleviate the unimodal bias. \citet{ramakrishnan18adversarial} employed adversarial training for VQA models. They posed training as an adversarial game between the VQA model and a question-only model, discouraging the VQA model from capturing language biases. \citet{lao21superficial} employed curriculum learning for VQA models. Their VQA model learns from the language modality first and then progressively learns multimodal reasoning from less-biased samples. RUBi \citep{cadene19RUBi} leverages a question-only model to dynamically adjust the loss in the training of VQA models. \citet{gat20regularize} proposed a new regularization term based on the functional entropy, which encourages the multimodal model to balance the contribution of each modality in training. \citet{wang20cvpr,peng22cvpr} inspect the contribution of each modality in the multimodal model and dynamically re-weight the loss \citep{wang20cvpr} or the gradient \citep{peng22cvpr} to balance the modalities during training. \citet{wu22greedy} inspect the speed at which the model learns from each modality and accelerate learning from underutilized modalities. \citet{du23laziness} introduced knowledge distillation to multimodel learning. They distill pretrained unimodal features into corresponding parts of the late fusion multimodal model to facilitate unimodal feature learning.

\section{Gradient Descent in Deep Multimodal Linear Networks}  \label{supp:grad}
We derive the gradient descent dynamics in multimodal deep linear networks with learning rate $\eta$.
In pre-fusion layers $1\leq l \leq L_f$, the gradient update is
\begin{subequations}
\begin{align}
\Delta \mW_\A^l &= 
- \eta \frac{\partial \Ls}{\partial \hat y} \frac{\partial \hat y}{\partial \mW_\A^l}  \nonumber
\\ &=
\eta \left( \prod_{j=L_f+1}^L \mW^j \prod_{i=l+1}^{L_f} \mW^i_\A \right)^\T 
(y - \hat y) 
\left( \prod_{i=1}^{l-1} \mW^i_\A \vx_\A \right)^\T  \nonumber
\\ &=
\eta \left( \prod_{j=L_f+1}^L \mW^j \prod_{i=l+1}^{L_f} \mW^i_\A \right)^\T
\left( y - \mW^{\tot}_\A \vx_\A - \mW^{\tot}_\B \vx_\B \right) \vx_\A^\T
\left( \prod_{i=1}^{l-1} \mW^i_\A \right)^\T  \nonumber
\\ &=
\eta \left( \prod_{j=L_f+1}^L \mW^j \prod_{i=l+1}^{L_f} \mW^i_\A \right)^\T 
\left( \mSigma_{y\vx_\A} - \mW^{\tot}_\A \mSigma_\A - \mW^{\tot}_\B \mSigma_{\B \A} \right)
\left( \prod_{i=1}^{l-1} \mW^i_\A \right)^\T
\\
\Delta \mW_\B^l &= \left( \prod_{j=L_f+1}^L \mW^j \prod_{i=l+1}^{L_f} \mW^i_\B \right)^\T 
\left( \mSigma_{y\vx_\B} -  \mW^{\tot}_\A \mSigma_{\B \A} -  \mW^{\tot}_\B \mSigma_\B \right)
\left( \prod_{i=1}^{l-1} \mW^i_\B \right)^\T 
\end{align} \label{eq:DeltaWBWA}
\end{subequations} 
In post-fusion layers $L_f + 1\leq l \leq L$,
\begin{align}
\Delta \mW^l &= 
- \eta \frac{\partial \Ls}{\partial \hat y} \frac{\partial \hat y}{\partial \mW^l}  \nonumber
\\ &=
\eta \left( \prod_{j=l+1}^L \mW^j \right)^\T 
(y - \hat y) 
\left( \prod_{j=L_f+1}^{l-1} \mW^j \prod_{i=1}^{L_f} \mW^i_\A \vx_\A \right)^\T 
  \nonumber \\
&+ \eta \left( \prod_{j=l+1}^L \mW^j \right)^\T 
(y - \hat y) 
\left( \prod_{j=L_f+1}^{l-1} \mW^j \prod_{i=1}^{L_f} \mW^i_\B \vx_\B \right)^\T  \nonumber
\\ &=
\eta \left( \prod_{j=l+1}^L \mW^j \right)^\T 
\left( y - \mW^{\tot}_\A \vx_\A - \mW^{\tot}_\B \vx_\B \right) \vx_\A^\T
\left( \prod_{j=L_f+1}^{l-1} \mW^j \prod_{i=1}^{L_f} \mW^i_\A \right)^\T 
  \nonumber \\
&+ \eta \left( \prod_{j=l+1}^L \mW^j \right)^\T 
\left( y - \mW^{\tot}_\A \vx_\A - \mW^{\tot}_\B \vx_\B \right) \vx_\B^\T
\left( \prod_{j=L_f+1}^{l-1} \mW^j \prod_{i=1}^{L_f} \mW^i_\B \right)^\T  \nonumber
\\ &=
\eta \left( \prod_{j=l+1}^L \mW^j \right)^\T 
\left( \mSigma_{y\vx_\A} -  \mW^{\tot}_\A \mSigma_\A -  \mW^{\tot}_\B \mSigma_{\B \A} \right)
\left( \prod_{j=L_f+1}^{l-1} \mW^j \prod_{i=1}^{L_f} \mW^i_\A \vx_\A \right)^\T 
  \nonumber \\
&+ \eta \left( \prod_{j=l+1}^L \mW^j \right)^\T 
\left( \mSigma_{y\vx_\B} - \mW^{\tot}_\A \mSigma_{\A \B} - \mW^{\tot}_\B \mSigma_\B \right)
\left( \prod_{j=L_f+1}^{l-1} \mW^j \prod_{i=1}^{L_f} \mW^i_\B \vx_\B \right)^\T
\label{eq:DeltaW}
\end{align}
In the limit of small learning rate, the difference equations \cref{eq:DeltaWBWA,eq:DeltaW} are well approximated by the differential equations \cref{eq:dWAdWB,eq:dW} in the main text.

\newpage
\section{Two-Layer Early Fusion Linear Network}  \label{supp:early-fixedpoint}
A two-layer early fusion linear network is described as $\hat y = \mW^2 \mW^1 \vx$. The gradient descent dynamics are
\begin{align} \label{eq:dW-early-2L}
\tau \dot \mW^1 = {\mW^2}^\T (\mSigma_{y\vx} - \mW^2 \mW^1 \mSigma )
, \quad
\tau \dot \mW^2 = (\mSigma_{y\vx} - \mW^2 \mW^1 \mSigma ) {\mW^1}^\T .
\end{align}
By setting to the gradients to zero, we find that there are two manifolds of fixed points:
\begin{subequations}
\begin{align}
\mW^2 = \vzero, \, \mW^1 = \vzero 
\quad &\Rightarrow \quad
\gM_0 = \{ \mW | \mW=\vzero \}  , \\
\mSigma_{y\vx} - \mW^2 \mW^1 \mSigma = \vzero
\quad &\Rightarrow \quad
\gM_* = \{ \mW | \mW^{\tot} = \mSigma_{y\vx} \mSigma^{-1} \}  .
\end{align}
\end{subequations}
We plot $\gM_0, \gM_*$ in \cref{fig:early_phase_portrait} for a scalar case introduced in \cref{fig:lin_early_rho0_2L}.

\section{Two-Layer Late Fusion Linear Network}
A two-layer late fusion linear network is described as $\hat y = \mW_\A^2 \mW_\A^1 \vx_\A + \mW_\B^2 \mW_\B^1 \vx_\B$. The gradient descent dynamics are
\begin{subequations}
\begin{align}
\tau \dot \mW_\A^1 &= {\mW_\A^2}^\T (\mSigma_{y\vx_\A} - \mW_\A^2 \mW_\A^1 \mSigma_\A - \mW_\B^2 \mW_\B^1 \mSigma_{\B \A})  ,  \\
\tau \dot \mW_\A^2 &= (\mSigma_{y\vx_\A} - \mW_\A^2 \mW_\A^1 \mSigma_\A - \mW_\B^2 \mW_\B^1 \mSigma_{\B \A}) {\mW_\A^1}^\T  ,  \\
\tau \dot \mW_\B^1 &= {\mW_\B^2}^\T (\mSigma_{y\vx_\B} - \mW_\A^2 \mW_\A^1 \mSigma_{\A \B} - \mW_\B^2 \mW_\B^1 \mSigma_\B)  ,  \\
\tau \dot \mW_\B^2 &= (\mSigma_{y\vx_\B} - \mW_\A^2 \mW_\A^1 \mSigma_{\A \B} - \mW_\B^2 \mW_\B^1 \mSigma_\B) {\mW_\B^1}^\T  .
\end{align}  \label{eq:late-grad-2L}%
\end{subequations}

\subsection{Fixed Points and Saddles}  \label{supp:late-fixedpoint}
By setting the gradients in \cref{eq:late-grad-2L} to zero, we find that the two manifolds of fixed points in \cref{eq:fixedpoint} exist in two-layer late fusion linear networks as well.
\begin{subequations}
\begin{align}
\begin{cases}
\mW_\A^2 = \vzero, \, \mW_\A^1 = \vzero  \\
\mW_\B^2 = \vzero, \, \mW_\B^1 = \vzero
\end{cases}
&\Rightarrow \quad
\gM_0 = \{ \mW | \mW=\vzero \}  , \\
\begin{cases}
\mSigma_{y\vx_\A} - \mW_\A^2 \mW_\A^1 \mSigma_\A - \mW_\B^2 \mW_\B^1 \mSigma_{\B \A} = \vzero  \\
\mSigma_{y\vx_\B} - \mW_\A^2 \mW_\A^1 \mSigma_{\A \B} - \mW_\B^2 \mW_\B^1 \mSigma_\B = \vzero
\end{cases}
&\Rightarrow \quad
\gM_* = \{ \mW | \mW^{\tot} = \mSigma_{y\vx} \mSigma^{-1} \}  .
\end{align}
\end{subequations}
In addition, there are two manifolds of saddles:
\begin{subequations}
\begin{align}
\begin{cases}
\mSigma_{y\vx_\A} - \mW_\A^2 \mW_\A^1 \mSigma_\A = \vzero  \\
\mW_\B^2 = \vzero, \, \mW_\B^1 = \vzero
\end{cases}
&\Rightarrow \quad
\gM_\A = \{ \mW | \mW^{\tot}_\A=\mSigma_{y \vx_\A} {\mSigma_\A}^{-1}, \mW^{\tot}_\B = \vzero \}  , \\
\begin{cases}
\mW_\A^2 = \vzero, \, \mW_\A^1 = \vzero  \\
\mSigma_{y\vx_\B} - \mW_\B^2 \mW_\B^1 \mSigma_\B = \vzero
\end{cases}
&\Rightarrow \quad
\gM_\B = \{ \mW | \mW^{\tot}_\A=\vzero, \mW^{\tot}_\B = \mSigma_{y \vx_\B} {\mSigma_\B}^{-1} \}  .
\end{align}
\end{subequations}
We plot the four manifolds $\gM_0, \gM_*, \gM_\A, \gM_\B$ in \cref{fig:late_phase_portrait} for a scalar case introduced in \cref{fig:lin_late_rho0_2L}.

\subsection{A Solvable Simple Case}
If the two modalities are not correlated ($\mSigma_{\A\B}=\vzero$) and have white covariance ($\mSigma_\A = \sigma_\A^2 \mI, \mSigma_\B = \sigma_\B^2 \mI$) the dynamics is equivalent to two separately trained unimodal two-layer linear networks with whitened input, whose solution has been derived in \citet{saxe13exact}. The time course solutions of total weights are
\begin{subequations}
\begin{align}
\mW_\A^{\tot} (t) &= u_\A (t) \sigma_\A^{-2} \mSigma_{y \vx_\A}
, \, u_\A (t) = \left[ \left( \frac{\| \mSigma_{y \vx_\A} \|}{\sigma_\A^2 u_\A (0)} -1 \right) e^{- 2 \| \mSigma_{y \vx_\A} \| \frac{t}{\tau}} + 1 \right]^{-1}  ,\\
\mW_\B^{\tot} (t) &= u_\B (t) \sigma_\B^{-2} \mSigma_{y \vx_\B}
, \, u_\B (t) = \left[ \left( \frac{\| \mSigma_{y \vx_\B} \|}{\sigma_\B^2 u_\B (0)} -1 \right) e^{- 2 \| \mSigma_{y \vx_\B} \| \frac{t}{\tau}} + 1 \right]^{-1}  .
\end{align} \label{eq:exact_sol}%
\end{subequations}
The total weights for both modalities only evolve in scale along the pseudo-inverse solution direction. The scale variables $u_\A (t), u_\B (t)$ both go through the sigmoidal growth while modality A grows approximately $\| \mSigma_{y \vx_\A} \| / \| \mSigma_{y \vx_\B} \|$ times faster.

\subsection{General Cases} \label{supp:time-2L}
In the general case of having arbitrary correlation matrices, the analytical solution in \cref{eq:exact_sol} does not hold since the total weights not only grow in one fixed direction but also rotate. We then focus on the early phase dynamics to compute the duration of the unimodal phase. 
\subsubsection{Which Modality is Learned First and When?}  \label{supp:prioritization}
We assume that the modality learned first is modality A and specify the time $t_\A$ and the condition for prioritizing modality A in the following. During time $0$ to $t_\A$, the total weights of both modalities have not moved much away from their small initialization near zero. Hence, the dynamics from time $0$ to $t_\A$ in leading order approximation of the initialization are
\begin{subequations}
\begin{align}
\tau \dot \mW_\A^1 = {\mW_\A^2}^\T \mSigma_{y\vx_\A} , \,
\tau \dot \mW_\A^2 = \mSigma_{y\vx_\A} {\mW_\A^1}^\T ; 
\\
\tau \dot \mW_\B^1 = {\mW_\B^2}^\T \mSigma_{y\vx_\B} , \,
\tau \dot \mW_\B^2 = \mSigma_{y\vx_\B} {\mW_\B^1}^\T .  \label{eq:early_uB}
\end{align}
\end{subequations}
The first-layer weights $\mW^1_\A, \mW^1_\B$ align with $\mSigma_{y\vx_\A}, \mSigma_{y\vx_\B}$ respectively in the early phase when their scale has not grown appreciably \citep{cengiz22silent}.
We have the balancing property \citep{du18autobalance,ji18align} between the two layers of modality A 
\begin{align}  \label{eq:def-uA}
\mW^1_\A {\mW^1_\A}^\T = {\mW^2_\A}^\T \mW^2_\A
\quad \Rightarrow \quad
\|\mW_\A^1\|_{\mathrm F} = \|\mW_\A^2\| \overset{\text{def}}{=} u_\A
.
\end{align}
We conduct the following change of variable
\begin{align}
\mW^1_\A = u_\A(t) \vr_\A^1 \frac{\mSigma_{y\vx_\A}}{\|\mSigma_{y\vx_\A}\|}
, \,
\mW^2_\A = u_\A(t) {\vr_\A^1}^\T  ,
\end{align}
where $\vr^1_\A$ is a fixed unit norm column vector representing the freedom in the hidden layer and $u_\A$ is a scalar representing the norm of the two balancing layers \citep{advani20highd}. Substituting the variables, we can re-write the dynamics of $\mW_\A^2$ and obtain an ordinary differential equation about $u_\A$:
\begin{align}
\tau \dot u_\A {\vr_\A^1}^\T = \mSigma_{y\vx_\A} \frac{\mSigma_{y\vx_\A}^\T}{\|\mSigma_{y\vx_\A}\|} {\vr_\A^1}^\T u_\A
\quad \Rightarrow \quad
\tau \dot u_\A = \|\mSigma_{y\vx_\A}\| u_\A .
\end{align}
Separating variables and integrating both sides, we get
\begin{align}
t = \tau \|\mSigma_{y\vx_\A}\|^{-1} ( \ln u_\A - \ln u_0 )  ,
\end{align}
where $u_0$ denotes $u_0 = u_\A(0)$.
Because the initialization $u_0$ is very small compared to $\|\mSigma_{y\vx_\A}\|$, the time that $u_\A$ grows to be comparable with $\|\mSigma_{y\vx_\A}\|$ is
\begin{align} \label{eq:tA}
t_\A \approx \tau \|\mSigma_{y\vx_\A}\|^{-1} \ln \frac1{u_0} ,
\end{align}
From \cref{eq:tA}, we can infer that the condition for modality A to be learned first is that $\|\mSigma_{y\vx_\A}\| > \|\mSigma_{y\vx_\B}\|$.
\begin{align}
\tau \|\mSigma_{y\vx_\A}\|^{-1} \ln \frac1{u_0} < \tau \|\mSigma_{y\vx_\B}\|^{-1} \ln \frac1{u_0} 
\quad \Leftrightarrow \quad
\|\mSigma_{y\vx_\A}\| > \|\mSigma_{y\vx_\B}\|  .
\end{align}

\subsubsection{When is the Second Modality Learned?}
We next calculate the time $t_\B$ when modality B is learned. During time $0$ to $t_\A$, the weight dynamics of modality B takes the same form as modality A as described in \cref{eq:early_uB}.
Because the layers in the modality B branch are balanced and $\mW^1_\B$ aligns with the rank-one direction $\mSigma_{y\vx_\B}$, we again change variables and re-write the dynamics of $\mW_\B^2$, obtaining an ordinary differential equation about the norm of the two balanced layers $u_\B = \|\mW_\B^1\|_{\mathrm F} = \|\mW_\B^2\|$:
\begin{align}  \label{eq:ode-uB}
\tau \dot u_\B & = \|\mSigma_{y\vx_\B}\| u_\B
\quad \Rightarrow \quad
t = \tau \|\mSigma_{y\vx_\B}\|^{-1} ( \ln u_\B - \ln u_0 )
, \, t \in [0, t_\A).
\end{align}
We assume $u_\B(0)=u_0$, since the initialization is small and the number of hidden neurons in the modality B branch is of the same order as modality A.
We plug $t_\A$, obtained in \cref{eq:tA}, into \cref{eq:ode-uB} and get
\begin{align}  \label{eq:uBtA}
\ln u_\B (t_\A) = \left( 1 - \| \mSigma_{y \vx_\A} \|^{-1} \| \mSigma_{y \vx_\B}\| \right) \ln u_0 .
\end{align}
We then look into the dynamics during the unimodal phase from $t_\A$ to $t_\B$. During time $t_\A$ to $t_\B$, the weights of modality B are still small and negligible in leading ordering approximation. Meanwhile, the weights of modality A have grown to be $\mW^{\tot}_\A=\mSigma_{y \vx_\A} {\mSigma_\A}^{-1}$, which influences the dynamics of modality B when the cross-covariance $\mSigma_{\A \B} \neq \vzero$. Taking this into consideration, the dynamics of modality B during $t_\A$ to $t_\B$ is
\begin{align}
\tau \dot \mW_\B^1 = {\mW_\B^2}^\T \widetilde \mSigma_{y\vx_\B}  , \,
\tau \dot \mW_\B^2 = \widetilde \mSigma_{y\vx_\B} {\mW_\B^1}^\T , \,
t \in (t_\A, t_\B) ,
\end{align}
where $\widetilde \mSigma_{y\vx_\B} = \mSigma_{y\vx_\B} - \mSigma_{y \vx_\A} {\mSigma_\A}^{-1} \mSigma_{\A \B}$. 
The first-layer weights $\mW_\B^1$ rapidly rotate from $\mSigma_{y\vx_\B}$ to $\widetilde \mSigma_{y\vx_\B}$ at time $t_\A$ and continue to evolve along $\widetilde \mSigma_{y\vx_\B}$ during $t_\A$ to $t_\B$. Through the same manner of changing variables, we obtain the ordinary differential equation about $u_\B$ during $t_\A$ to $t_\B$:
\begin{align}
\tau \dot u_\B & = \left\|\widetilde \mSigma_{y\vx_\B} \right\| u_\B
\quad \Rightarrow \quad
t - t_\A = \tau \left\|\widetilde \mSigma_{y\vx_\B} \right\|^{-1} ( \ln u_\B - \ln u_\B (t_\A) )
, \, t \in (t_\A, t_\B).
\end{align}
Plugging in $u_\B (t_\A)$ obtained in \cref{eq:uBtA}, we get the time when $u_\B$ grows to be comparable with $\| \mSigma_{y\vx_\B} \|$:
\begin{align} \label{eq:tB}
t_\B \approx 
t_\A - \tau \left\| \widetilde \mSigma_{y\vx_\B} \right\|^{-1} \ln u_\B (t_\A)  \approx 
t_\A + \tau \frac{1 - \| \mSigma_{y \vx_\A} \|^{-1} \| \mSigma_{y \vx_\B}\|}{\left\| \widetilde \mSigma_{y\vx_\B} \right\|} \ln \frac1{u_0}  .
\end{align}
Dividing \cref{eq:tB} by \cref{eq:tA}, we obtain the time ratio \cref{eq:timeratio-2L} in the main text:
\begin{align*}
\frac{t_\B}{t_\A} \approx 1 + \frac{\| \mSigma_{y \vx_\A} \| - \| \mSigma_{y \vx_\B} \|} {\lnorm \mSigma_{y \vx_\B} - \mSigma_{y \vx_\A} {\mSigma_\A}^{-1} \mSigma_{\A \B} \rnorm}  .
\end{align*}

\begin{figure}[h]
\centering
\includegraphics[width=.27\linewidth]{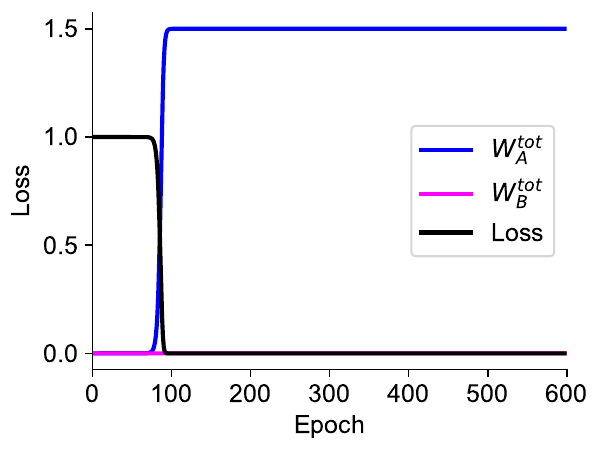}
\caption{Loss and weight trajectories in a two-layer late fusion network when modalities have collinearity. Inputs $\vx_\A$ and $\vx_\B$ are scalars with covariance matrix $\mSigma=[4,2;2,1]$. The target output is generated as $y=\vx_\A+\vx_\B$.}
\label{fig:collinear}
\hfill
\end{figure}

We validate \cref{eq:timeratio-2L} with numerical simulations in \cref{fig:timesweep_2L}. 
Note that if $\vx_\A$ and $\vx_\B$ have collinearity, the denominator in \cref{eq:timeratio-2L} is zero and the time ratio is $\infty$. As shown in \cref{fig:collinear}, the late fusion network learns to fit the output only with modality A and modality B will never be learned.

\subsection{Logistic Loss}
\begin{figure}[h]
\centering
\subfloat[Early fusion]
{\centering
\includegraphics[width=0.3\linewidth]{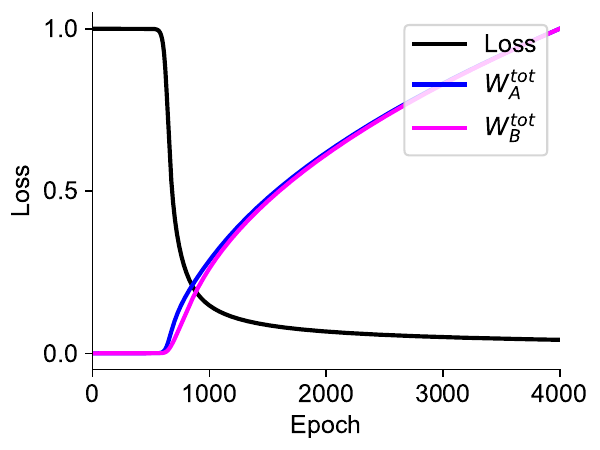}}
\subfloat[Late fusion]
{\centering
\includegraphics[width=0.3\linewidth]{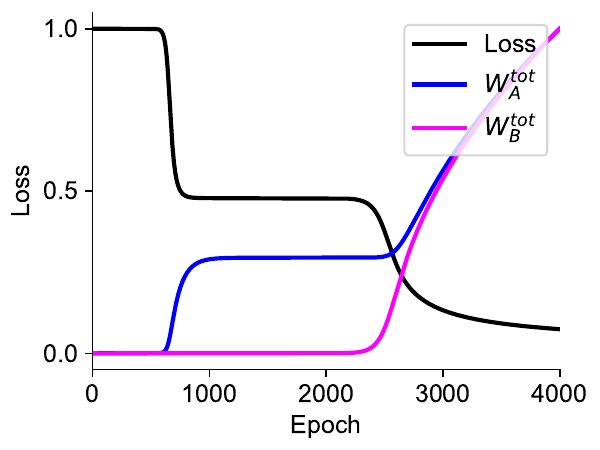}}
\hspace{2ex}
\subfloat[Time ratio in late fusion  \label{fig:timesweep_2L_logistic}]
{\centering
\includegraphics[width=0.27\linewidth]{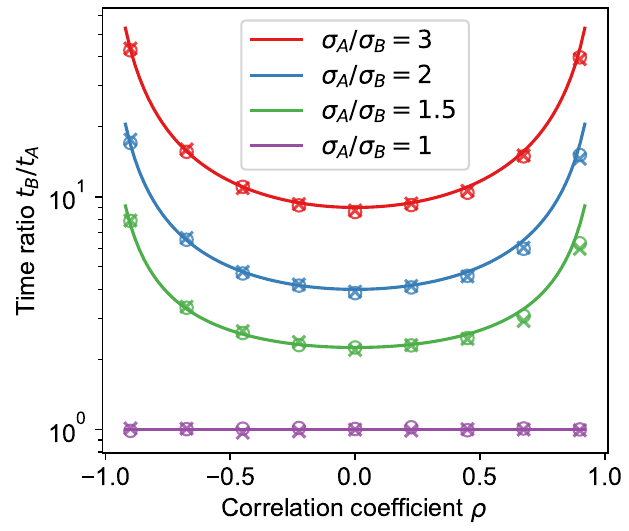}}
\caption{Two-layer early and late fusion linear networks trained with logistic loss. (a,b) Loss and weights trajectories. (c) Time ratio as in \cref{eq:timeratio-2L}.}
\label{fig:logistic}
\end{figure}

Our qualitative and quantitative results, derived with mean square error loss, carry over to multimodal deep linear network with logistic loss.
As shown in \cref{fig:logistic}, the early fusion linear network trained with logistic loss learns both modality almost simultaneously while the late fusion network learns two modalities at two separate times, which is similar to \cref{fig:lin_early_rho0_2L,fig:lin_late_rho0_2L}.
Note that weights diverge due to the nature of the logistic loss, which cannot attain zero with any finite weights and keeps decreasing for larger $\hat y$ values with the correct sign.

The time ratio in late fusion linear networks trained with logistic loss is the same as those trained with mean square error loss as demonstrated in \cref{fig:timesweep_2L_logistic}. This is because the first order approximation of the gradient descent dynamics in the early phase of training is the same for the two loss functions.

In the early phase of training, the network output $\hat y$ is close to 0 since we assume small initialization. Thus the Taylor expansion of the loss function around $\hat y=0$ is a good approximation in the early phase
\begin{align}
\ell_{\text{MSE}}(\hat y) &\equiv \frac12 (\hat y - y)^2
= \frac12 y^2 - y \hat y + O(\hat y^2)  ,
\\
\ell_{\text{LG}}(\hat y)
&\equiv \ln \left(1 + \exp(- y^\mu\hat y^\mu)\right)
= \ell_{\text{LG}}(0) + \ell_{\text{LG}}'(0) \hat y + O(\hat y^2)
= \ln2 -\frac 12 y \hat y + O(\hat y^2)  .
\end{align}
We can use the first-order expansion to write the gradient descent dynamics in the early phase
\begin{align}
\tau \dot \mW &= - \frac{d \ell_{\text{MSE}}}{d\hat y} \frac{d \hat y}{d\mW}
\approx y \frac{d \hat y}{d\mW}  ,
\\
\tau \dot \mW &= - \frac{d \ell_{\text{LG}}}{d\hat y} \frac{d \hat y}{d\mW}
\approx \frac12 y \frac{d \hat y}{d\mW}  ,
\end{align}
which are the same for square and logistic loss except a constant factor. Since the calculation of the time ratio mainly involves the early phase dynamics, the time ratio in late fusion linear networks remains the same whether trained with mean square error or logistic loss.

\section{Deep Intermediate \& Late Fusion Linear Networks}  \label{supp:time-deep}
\subsection{Balancing Properties}  \label{supp:balancing}
The full dynamics of deep intermediate and late fusion linear networks are given in \cref{eq:dWAdWB,eq:dW}. Between pre-fusion layers and between post-fusion layers, the standard balancing property \citep{du18autobalance,ji18align} holds true:
\begin{subequations}
\begin{align}
&\mW^1_\A {\mW^1_\A}^\T 
= \mW^2_\A {\mW^2_\A}^\T 
= \cdots
= \mW^{L_f}_\A {\mW^{L_f}_\A}^\T  , \\
&\mW^1_\B {\mW^1_\B}^\T 
= \mW^2_\B {\mW^2_\B}^\T 
= \cdots
= \mW^{L_f}_\B {\mW^{L_f}_\B}^\T  , \\
&\mW^{L_f + 1} {\mW^{L_f + 1}}^\T 
= \mW^{L_f + 2} {\mW^{L_f + 2}}^\T
= \cdots
= {\mW^L}^\T \mW^L  .
\end{align}  \label{eq:deep-balancing-old}%
\end{subequations}
Between a pre-fusion layer and a post-fusion layer, the balancing condition takes a slightly different form:
\begin{align}  \label{eq:deep-balancing-new}
\mW^{L_f}_\A {\mW^{L_f}_\A}^\T + \mW^{L_f}_\B {\mW^{L_f}_\B}^\T = \mW^{L_f + 1} {\mW^{L_f + 1}}^\T  .
\end{align}
Based on the standard balancing properties in \cref{eq:deep-balancing-old}, we have equal norm between pre-fusion layers and between post-fusion layers.
\begin{subequations}
\begin{align}
&\|\mW_\A^1\|_{\mathrm F} = \|\mW_\A^2\|_{\mathrm F} = \cdots = \|\mW_\A^{L_f}\|_{\mathrm F}
\overset{\text{def}}{=} u_\A  ,\\
&\|\mW_\B^1\|_{\mathrm F} = \|\mW_\B^2\|_{\mathrm F} = \cdots = \|\mW_\B^{L_f}\|_{\mathrm F}
\overset{\text{def}}{=} u_\B  ,\\
&\|\mW^{L_f+1}\|_{\mathrm F} = \|\mW^{L_f+2}\|_{\mathrm F} = \cdots = \|\mW^{L}\|
\overset{\text{def}}{=} u  .
\end{align}
\end{subequations}
Based on \cref{eq:deep-balancing-new}, we have the balancing property between the norm of a pre-fusion layer and a post-fusion layer.
\begin{align}  \label{eq:deep_uAuBu}
u_\A^2+u_\B^2 = u^2 .
\end{align}
In addition to the balancing norm, we infer from the balancing properties that post-fusion layers have rank-one structures as in standard linear networks \citep{ji18align,cengiz22silent}.
The pre-fusion layers are not necessarily rank-one due to the different balancing property in \cref{eq:deep-balancing-new}. However, guided by empirical observations, we make the ansatz that the weights in pre-fusion layers are also rank-one, which will enable us to conduct the change of variables similar to what we have done in \cref{supp:time-2L}.

\subsection{Time Ratio for $L_f \neq 2$}  \label{supp:time-deep-general}
\subsubsection{Which Modality is Learned First and When?}
We adopt the convention that the modality learned first is modality A. During time $0$ to $t_\A$, the dynamics in leading order approximation of the initialization are
\begin{subequations}
\begin{align}
\tau \dot \mW_\A^l &= \left( \prod_{j=L_f+1}^L \mW^j \prod_{i=l+1}^{L_f} \mW^i_\A \right)^\T 
\mSigma_{y\vx_\A}
\left( \prod_{i=1}^{l-1} \mW^i_\A \right)^\T 
,\,  1 \leq l \leq L_f  , \\
\tau \dot \mW_\B^l &= \left( \prod_{j=L_f+1}^L \mW^j \prod_{i=l+1}^{L_f} \mW^i_\B \right)^\T 
\mSigma_{y\vx_\B}
\left( \prod_{i=1}^{l-1} \mW^i_\B \right)^\T 
,\,  1 \leq l \leq L_f  , \\
\tau \dot \mW^l &= \left( \prod_{j=l+1}^L \mW^j \right)^\T 
\mSigma_{y\vx_\A}
\left( \prod_{j=L_f+1}^{l-1} \mW^j \prod_{i=1}^{L_f} \mW^i_\A \right)^\T 
  \nonumber \\
&+ \left( \prod_{j=l+1}^L \mW^j \right)^\T 
\mSigma_{y\vx_\B}
\left( \prod_{j=L_f+1}^{l-1} \mW^j \prod_{i=1}^{L_f} \mW^i_\B \right)^\T   
,\,  L_f+1 \leq l \leq L.
\end{align}  \label{eq:approx-dWAdWBdW}%
\end{subequations}

We make the following change of variables
\begin{subequations}
\begin{align}
\mW^1_\A &= u_\A(t) \vr^1_\A \frac{\mSigma_{y\vx_\A}}{\|\mSigma_{y\vx_\A}\|} , \, \mW^1_\B = u_\B(t) \vr^1_\B \frac{\mSigma_{y\vx_\B}}{\|\mSigma_{y\vx_\B}\|}  ; \\
\mW^i_\A &= u_\A(t) \vr^i_\A {\vr^{i-1}_\A}^\T , \, \mW^i_\B = u_\B(t) \vr^i_\B {\vr^{i-1}_\B}^\T , \, 2 \leq i \leq L_f-1  ; \\
\mW^{L_f}_\A &= u_\A(t) \vr^{L_f} {\vr^{L_f-1}_\A}^\T , \, \mW^{L_f}_\B = u_\B(t) \vr^{L_f} {\vr^{L_f-1}_\B}^\T  ; \\
\mW^j &= u(t) \vr^j {\vr^{j-1}}^\T , \, L_f+1 \leq j \leq L-1  ; \\
\mW^L &= u(t) {\vr^L}^\T  .
\end{align}  \label{eq:change-var-deep}%
\end{subequations}
where all $\vr$ are fixed unit norm column vectors representing the freedom in hidden layers. By substituing \cref{eq:change-var-deep} into \cref{eq:approx-dWAdWBdW}, we reduce the dynamics during time $0$ to $t_\A$ to a three-dimensional dynamical system about $u_\A, u_\B$ and $u$:
\begin{subequations}
\begin{align}
\label{eq:deep_uA}
\tau \dot u_\A &= \| \mSigma_{y\vx_\A} \| u_\A^{L_f-1} u^{L-L_f} ,  \\ 
\label{eq:deep_uB}
\tau \dot u_\B &= \| \mSigma_{y\vx_\B} \| u_\B^{L_f-1} u^{L-L_f} ,  \\
\tau \dot u &= \| \mSigma_{y\vx_\A} \| u_\A^{L_f} u^{L-L_f-1}  + \| \mSigma_{y\vx_\B} \| u_\B^{L_f} u^{L-L_f-1}  .
\end{align}
\end{subequations}
We divide \cref{eq:deep_uB} by \cref{eq:deep_uA} and reveal an equality between the two branches of a pre-fusion layer:
\begin{align} \label{eq:deep_uAuB}
\frac{d u_\B}{d u_\A} = \frac{\| \mSigma_{y\vx_\B} \| u_\B^{L_f-1}}{\| \mSigma_{y\vx_\A} \| u_\A^{L_f-1}} 
\quad \Rightarrow \quad
\frac{u_\B^{2-L_f} - u_0^{2-L_f}}{u_\A^{2-L_f} - u_0^{2-L_f}} = 
\frac{\| \mSigma_{y\vx_\B} \|}{\| \mSigma_{y\vx_\A} \|}
\overset{\text{def}}{=} k
,\, L_f \neq 2 ,
\end{align}
We assume $L_f \neq 2$ in this section and handle the case of $L_f=2$ in \cref{supp:time-deep-2Lf}.
By utilizing two equality properties in \cref{eq:deep_uAuB} and \cref{eq:deep_uAuBu}, we reduce \cref{eq:deep_uA} to an ordinary differential equation about $u_\A$
\begin{align}  \label{eq:deep_ode-uA}
\tau \dot u_\A = \| \mSigma_{y\vx_\A} \| u_\A^{L_f-1} \left[ u_\A^2 + \left( k u_\A^{2-L_f} + \left(1-k \right) u_0^{2-L_f} \right)^{\frac{2}{2-L_f}} \right] ^{\frac{L-L_f}{2}}  .
\end{align}
The ordinary differential equation in \cref{eq:deep_ode-uA} is separable despite being cumbersome. By separating variables and integrating both sides, we can write time $t$ as a function of $u_\A$:
\begin{align}
t = \int_{u_0}^{u_\A} x^{1-L_f} \left[ x^2 + \left( k x^{2-L_f} + \left(1-k \right) u_0^{2-L_f} \right)^{\frac{2}{2-L_f}} \right] ^{\frac{L_f-L}{2}} dx  .
\end{align}
The time when $u_\A$ grows to be comparable with $\| \mSigma_{y\vx_\A} \|$, for instance $\| \mSigma_{y\vx_\A} \|/2$, is
\begin{align}  \label{eq:deep-tA}
t_\A &\approx \tau \| \mSigma_{y\vx_\A} \|^{-1} \int_{u_0} ^{\| \mSigma_{y\vx_\A} \|/2}
u_\A^{1-L_f} \left[ u_\A^2 + \left( k u_\A^{2-L_f} + \left(1-k \right) u_0^{2-L_f} \right)^{\frac{2}{2-L_f}} \right] ^{\frac{L_f-L}{2}} du_\A  \nonumber  \\
&\approx \tau \| \mSigma_{y\vx_\A} \|^{-1} u_0^{2-L} \int_{1} ^ \infty x^{1-L} \left[ 1 + \left( k + \left(1-k\right) x^{L_f-2} \right)^{\frac2{2-L_f}} \right]^{\frac{L_f-L}2} dx  .
\end{align}
From \cref{eq:deep-tA}, we find that the condition for modality A to be learned first in deep intermediate and late fusion linear networks is the same as that for two-layer late fuison linear networks:
\begin{align}
\tau \|\mSigma_{y\vx_\A}\|^{-1} u_0^{2-L_f} I(L, L_f) < \tau \|\mSigma_{y\vx_\B}\|^{-1} u_0^{2-L_f} I(L, L_f) 
\quad \Leftrightarrow \quad
\frac{\|\mSigma_{y\vx_\B}\|}{\|\mSigma_{y\vx_\A}\|} \equiv k \in (0,1) ,
\end{align}
where we use $I(L, L_f)$ to denote the integral
\begin{align*}
I(L, L_f) &= \int_{1} ^ \infty x^{1-L} \left[ 1 + \left( k + \left(1-k \right) x^{L_f-2} \right)^{\frac2{2-L_f}} \right]^{\frac{L_f-L}2} dx
\, .
\end{align*}

\subsubsection{When is the Second Modality Learned?}
We next compute the time $t_\B$ when modality B is learned. During time $t_\A$ to $t_\B$, the network is in manifold $\gM_\A$ where 
\begin{align}
\mW^{\tot}_\A=\mSigma_{y \vx_\A} {\mSigma_\A}^{-1}
\quad \Rightarrow \quad
u_\A = u = \lnorm \mSigma_{y \vx_\A} {\mSigma_\A}^{-1} \rnorm^{\frac1L}
, \,
t \in (t_\A, t_\B)  .
\end{align}
We plug $u_\A(t_\A)$, which is very large compared to $u_0$, into \cref{eq:deep_uAuB} and obtain $u_\B(t_\A)$
\begin{align}  \label{eq:uBtA-deep}
u_\B(t_\A) = \left[ u_\A(t_\A)^{2-L_f} + (1-k) u_0^{2-L_f} \right] ^{\frac1{2-L_f}}
\approx \left(1-k \right)^{\frac1{2-L_f}} u_0  .
\end{align}
We then look into the dynamics during the unimodal phase from $t_\A$ to $t_\B$. Substituting $\mW^{\tot}_\A=\mSigma_{y \vx_\A} {\mSigma_\A}^{-1}$ into \cref{eq:dWB}, we get
\begin{align}
\tau \dot \mW_\B^l &= \left( \prod_{j=L_f+1}^L \mW^j \prod_{i=l+1}^{L_f} \mW^i_\B \right)^\T 
\widetilde \mSigma_{y\vx_\B}
\left( \prod_{i=1}^{l-1} \mW^i_\B \right)^\T
, \,  1 \leq l \leq L_f  ,
\end{align}
where $\widetilde \mSigma_{y\vx_\B} = \mSigma_{y\vx_\B} - \mSigma_{y \vx_\A} {\mSigma_\A}^{-1} \mSigma_{\A \B}$. The first-layer weights $\mW^1_\B$ rapidly rotate from $\mSigma_{y\vx_\B}$ to $\widetilde \mSigma_{y\vx_\B}$ at time $t_\A$ and continues to evolve along $\widetilde \mSigma_{y\vx_\B}$ during $t_\A$ to $t_\B$.
Through the same manner of changing variables, we obtain an ordinary differential equation for $u_\B$ during $t_\A$ to $t_\B$:
\begin{align}
\begin{split}
\tau \dot u_\B &= \lnorm \mSigma_{y \vx_\A} {\mSigma_\A}^{-1} \rnorm^{1-\frac{L_f}L} \|\widetilde \mSigma_{y\vx_\B}\| u_\B^{L_f-1}
\\ \Rightarrow \quad
t - t_\A &= \tau \lnorm \mSigma_{y \vx_\A} {\mSigma_\A}^{-1} \rnorm^{\frac{L_f}L-1} \|\widetilde \mSigma_{y\vx_\B}\|^{-1}
\frac{u_\B^{2-L_f} - u_\B(t_\A)^{2-L_f}}{2-L_f}
, \, t \in (t_\A, t_\B).
\end{split}
\end{align}
Plugging in $u_\B(t_\A)$ obtained in \cref{eq:uBtA-deep}, we get the time $t_\B$:
\begin{align}
\begin{split}
\label{eq:deep-tB}
t_\B &= 
t_\A + \tau \frac{u_\B(t_\B)^{2-L_f} - u_\B (t_\A)^{2-L_f}}{(2-L_f)\lnorm \mSigma_{y \vx_\A} {\mSigma_\A}^{-1} \rnorm^{1-\frac{L_f}L} \|\widetilde \mSigma_{y\vx_\B}\|}  \\
&\approx t_\A + \tau \frac{- u_\B (t_\A)^{2-L_f}}{(2-L_f)\lnorm \mSigma_{y \vx_\A} {\mSigma_\A}^{-1} \rnorm^{1-\frac{L_f}L} \|\widetilde \mSigma_{y\vx_\B}\|}  \\
&\approx t_\A + \tau \frac{1 - \| \mSigma_{y \vx_\A} \|^{-1} \| \mSigma_{y \vx_\B}\|}{(L_f-2)\lnorm \mSigma_{y \vx_\A} {\mSigma_\A}^{-1} \rnorm^{1-\frac{L_f}L} \|\widetilde \mSigma_{y\vx_\B}\|} u_0^{2-L_f}  .
\end{split}
\end{align}
Dividing \cref{eq:deep-tB} by \cref{eq:deep-tA}, we arrive at the time ratio \cref{eq:deep-late-timeratio} in the main text. For intermediate and late fusion linear networks $2<L_f\leq L$, the time ratio is
\begin{align*}
\frac{t_\B}{t_\A} &= 1 + \frac{(\| \mSigma_{y \vx_\A} \|-\| \mSigma_{y \vx_\B} \|) u_0^{L-L_f}} {(L_f-2)\lnorm \mSigma_{y \vx_\A} {\mSigma_\A}^{-1} \rnorm^{1-\frac{L_f}L} \lnorm \mSigma_{y \vx_\B} - \mSigma_{y \vx_\A} {\mSigma_\A}^{-1} \mSigma_{\A \B} \rnorm} I(L, L_f)^{-1}  ,
\end{align*}
where the integral $I(L,L_f)$ has been defined in \cref{eq:integral}.

\subsection{Time Ratio for $L_f = 2$}  \label{supp:time-deep-2Lf}
When the fusion layer is the second layer $L_f = 2$, the equality in \cref{eq:deep_uAuB} takes following form:
\begin{align}
\frac{d u_\A}{d u_\B} = \frac{\| \mSigma_{y\vx_\A} \| u_\A}{\| \mSigma_{y\vx_\B} \| u_\B} 
\quad \Rightarrow \quad
\frac{\ln u_\A - \ln u_0}{\ln u_\B - \ln u_0} = 
\frac{\| \mSigma_{y\vx_\A} \|}{\| \mSigma_{y\vx_\B} \|}  .
\end{align}
Consequently, fusion at the second layer is a special case with slightly different expressions for the times. We follow the same procedure as \cref{supp:time-deep-general} and obtain the times for $2 = L_f < L$:
\begin{subequations}
\begin{align}
t_\A &\approx \tau \frac{u_0^{2-L}}{\| \mSigma_{y\vx_\A} \|} \int_{1} ^ \infty x^{-1} \left( x^2 + x^{2k} \right)^{1-\frac{L}2} dx  ,
\\
t_\B &\approx t_\A + \tau \frac{1 - \| \mSigma_{y \vx_\A} \|^{-1} \| \mSigma_{y \vx_\B}\|} {\lnorm \mSigma_{y \vx_\B} - \mSigma_{y \vx_\A} {\mSigma_\A}^{-1} \mSigma_{\A \B} \rnorm} \lnorm \mSigma_{y \vx_\A} {\mSigma_\A}^{-1} \rnorm^{\frac{2}{L}-1} \ln \frac1{u_0}  .
\end{align}
\end{subequations}
The time ratio is
\begin{align}
\frac{t_\B}{t_\A} = 1 + \frac{(\| \mSigma_{y \vx_\A} \|-\| \mSigma_{y \vx_\B} \|) u_0^{L-2} \ln \frac1{u_0}} {(L_f-2)\lnorm \mSigma_{y \vx_\B} - \mSigma_{y \vx_\A} {\mSigma_\A}^{-1} \mSigma_{\A \B} \rnorm \lnorm \mSigma_{y \vx_\A} {\mSigma_\A}^{-1} \rnorm^{1-\frac2L}} I(L, 2)^{-1}  ,
\end{align}
where the integral is given by
\begin{align}
I(L, 2) = 
\int_{1} ^ \infty x^{-1} \left( x^2 + x^{2k} \right)^{1-\frac{L}2} dx .
\end{align}

\subsection{Time Ratio for Unequal Depth}
Our time ratio calculations can be carried out for intermediate fusion linear networks with unequal depth between modalities. Consider a intermediate fusion linear network with $L_c$ post-fusion layers, $L_\A$ pre-fusion layers for the modality A branch, and $L_\B$ pre-fusion layers for the modality B branch. Assuming $L_\A, L_\B>2$ and modality A is learned first, the time ratio is
\begin{align}
\frac{t_\B}{t_\A} = 1+
\frac{\frac{u_0^{L_c+L_\A-L_\B}}{L_\B-2}\| \mSigma_{yx_\A} \| - \frac{u_0^{L_c}}{L_\A-2} \| \mSigma_{yx_\B} \| }
{\lnorm \mSigma_{yx_\A} \mSigma_\A^{-1} \rnorm^{\frac{L_c}{L_\A+L_c}} \lnorm \mSigma_{yx_\B} - \mSigma_{yx_\A} \mSigma_\A^{-1} \mSigma_{\A\B} \rnorm}
I(L_\A, L_\B, L_c)^{-1} ,
\end{align}
where the integral is given by
\begin{align}
I(L_\A, L_\B, L_c) =
\int_1^\infty x^{1-L_\A} \left[
x^2 + \left( \frac{(L_\B-2)\| \boldsymbol \Sigma_{yx_\B} \|}{(L_\A-2)\| \boldsymbol \Sigma_{yx_\A} \|} u_0^{L_\B-L_\A} \left(x^{2-L_\A}-1\right)+1
\right)^{\frac2{2-L_\B}}
\right]^{-\frac{L_c}2} dx .
\end{align}
As a sanity check, setting $L_\A=L_\B$ gives us back the same expression as \cref{eq:deep-late-timeratio}.

\section{Feature Evolution in Multimodal Linear Networks}
\begin{figure}
\centering
\subfloat[Loss and weights]
{\centering
\includegraphics[width=0.3\linewidth]{fig/lin_early.pdf}}
\subfloat[First layer weights evolution in early fusion linear network  \label{fig:lin_early_feat}]
{\centering
\includegraphics[width=0.7\linewidth]{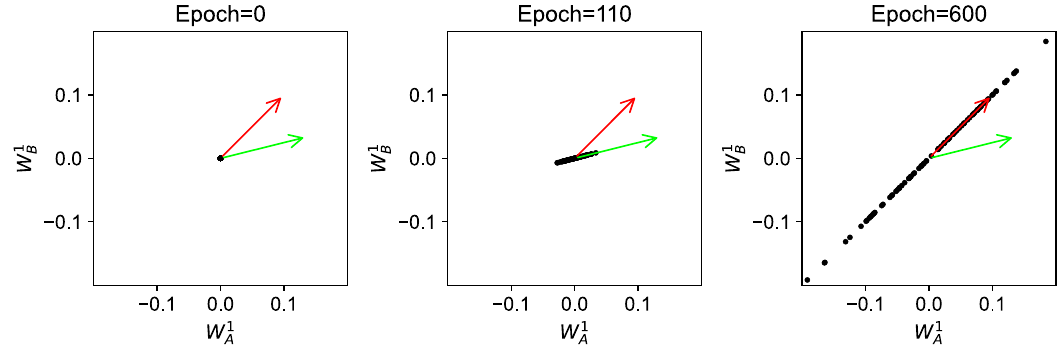}}
\\[2ex]
\subfloat[Loss and weights]
{\centering
\includegraphics[width=0.3\linewidth]{fig/lin_late.pdf}}
\subfloat[First layer weights evolution in late fusion linear network \label{fig:lin_late_feat}]
{\centering
\includegraphics[width=0.7\linewidth]{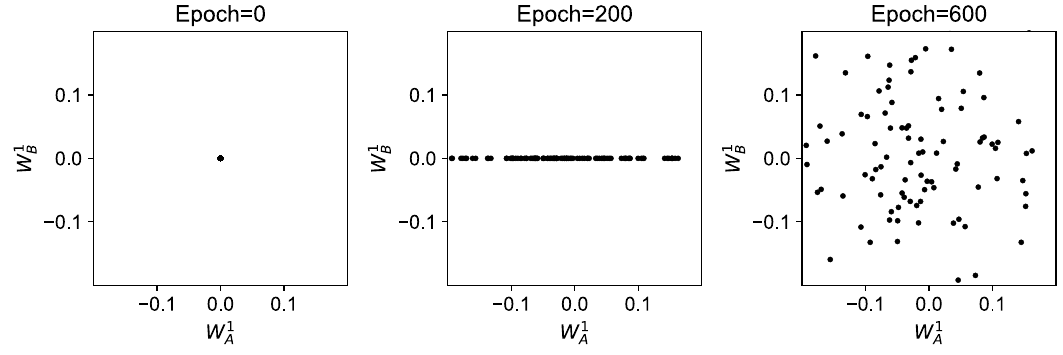}}
\caption{Feature evolution in two-layer early fusion and late fusion linear networks.
We plot the feature evolution, which are first-layer weights in linear networks, corresponding to dynamics in \cref{fig:lin_early_rho0_2L,fig:lin_late_rho0_2L}.
(a) Same as \cref{fig:lin_early_rho0_2L}. (b) First-layer weights at initialization, during training, and at convergence in panel a. The green arrow denotes the direction of $\mSigma_{y\vx}$. The red arrow denotes the direction of $\mSigma_{y\vx} \mSigma^{-1}$.
(c) Same as \cref{fig:lin_late_rho0_2L}. (d) First-layer weights at initialization, during training, and at convergence in panel c. 
Complementary videos can be found on our website.}
\label{fig:lin_feat_2L}
\end{figure}

We compare the feature evolution in two-layer early fusion and late fusion linear networks to gain a more detailed understanding of their different learning dynamics. As studied by \citet{cengiz22silent}, features of linear networks lie in the first-layer weights. We thus plot the first-layer weights in multimodal linear networks at different times in training to inspect the feature evolution.

\subsection{Feature Evolution in Early Fusion Linear Networks}
In early fusion linear networks with small initialization, the balancing property $\mW^1 {\mW^1}^\T = {\mW^2}^\T \mW^2$ holds true throughout training, which implies $\mW^1$ is rank-one throughout training \citep{du18autobalance,ji18align}. Specifically, $\mW^1$ initially aligns with $\mSigma_{y\vx}$ during the plateau and eventually aligns with $\mSigma_{y\vx} \mSigma^{-1}$ after $\mW^1$ have grown in scale and rotated during the brief transition stage. We illustrate this process with \cref{fig:lin_early_feat}. In deep early fusion linear networks, $\mW^1$ behaves qualitatively the same.

\subsection{Feature Evolution in Intermediate and Late Fusion Linear Networks}
In intermediate and late fusion linear networks, the balancing property takes a different form as given in \cref{eq:deep-balancing-old,eq:deep-balancing-new}. Thus the first-layer weights are not constrained to be rank-one. Specifically, $\mW^1_\A$ grows during the first transition stage while $\mW^1_\B$ remains close to small initialization. After a unimodal phase, $\mW^1_\B$ starts to grow during the second transition stage while $\mW^1_\A$ stays unchanged, shrinks in scale, or expands in scale depending on modality $\B$ has zero, positive, or negative correlation with modality $\A$. We illustrate this process with \cref{fig:lin_late_feat}.
In deep late and intermediate fusion linear networks, $\mW^1$ behaves qualitatively the same as in two-layer late fusion linear networks.

\section{Superficial Modality Preference}
We impose that modality A contributes less to the output, meaning the loss would decrease more if the network visited a saddle in $\gM_\B$ instead of $\gM_\A$
as illustrated in \cref{fig:superficial-ex}:
\begin{align}  \label{eq:superficial-ls-ineq}
\Ls (\gM_\A)  > \Ls (\gM_\B)
\end{align}
Substituting in the saddles defined in \cref{eq:saddles}, we expand and simplify the two mean square losses:
\begin{subequations}
\begin{align}
\Ls (\gM_\A) &=
\left\langle (y - \mSigma_{y \vx_\A} {\mSigma_\A}^{-1} \vx_\A)^2 \right\rangle   \nonumber \\
&=  \left\langle y^2 \right\rangle - 2 \mSigma_{y \vx_\A} {\mSigma_\A}^{-1} \langle y\vx_\A \rangle + \mSigma_{y \vx_\A} {\mSigma_\A}^{-1} \left\langle \vx_\A \vx_\A^\T \right\rangle {\mSigma_\A}^{-1} \mSigma_{y \vx_\A}^\T  \nonumber \\
&=  \left\langle y^2 \right\rangle - \mSigma_{y \vx_\A} {\mSigma_\A}^{-1} \mSigma_{y \vx_\A}^\T
,\\
\Ls (\gM_\B) &=
\left\langle (y - \mSigma_{y \vx_\B} {\mSigma_\B}^{-1} \vx_\B)^2  \right\rangle  \nonumber \\
&= \left\langle y^2 \right\rangle - 2 \mSigma_{y \vx_\B} {\mSigma_\A}^{-1} \langle y\vx_\B \rangle + \mSigma_{y \vx_\B} {\mSigma_\B}^{-1} \left\langle \vx_\B \vx_\B^\T \right\rangle {\mSigma_\B}^{-1} \mSigma_{y \vx_\B}^\T  \nonumber \\
&= \left\langle y^2 \right\rangle - \mSigma_{y \vx_\B} {\mSigma_\B}^{-1} \mSigma_{y \vx_\B}^\T  .
\end{align}  \label{eq:ls-expand}%
\end{subequations}
Plugging \cref{eq:ls-expand} into \cref{eq:superficial-ls-ineq} gives us
\begin{align*}
\mSigma_{y \vx_\A} {\mSigma_\A}^{-1} \mSigma_{y \vx_\A}^\T &< \mSigma_{y \vx_\B} {\mSigma_\B}^{-1} \mSigma_{y \vx_\B}^\T  .
\end{align*}
Since we also assume modality A is learned first, thus $\| \mSigma_{y \vx_\A} \| > \| \mSigma_{y \vx_\B} \|$.
We thereby arrive at the two inequality conditions \cref{eq:superficial} in the main text.

\section{Two-Layer Multimodal ReLU Networks}

\subsection{ReLU Networks with Linear Task}
\begin{figure}[h]
\centering
\subfloat[Early fusion]
{\centering
\includegraphics[width=0.3\linewidth]{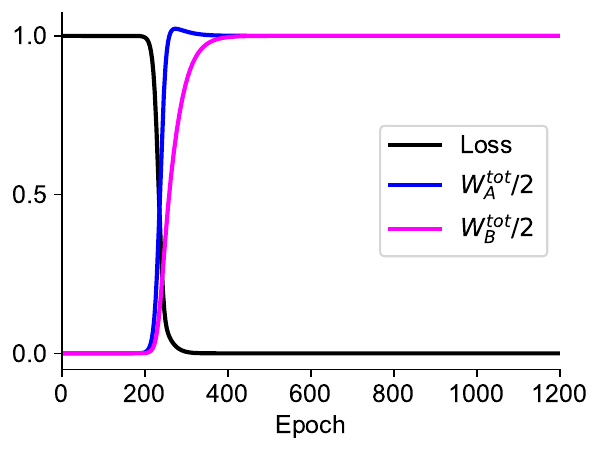}}
\hspace{2ex}
\subfloat[Late fusion]
{\centering
\includegraphics[width=0.3\linewidth]{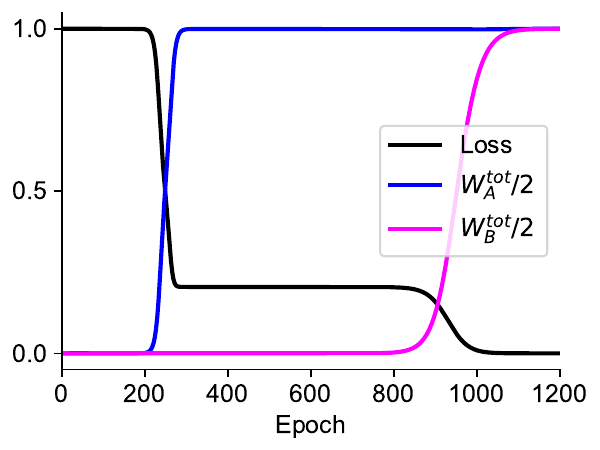}}
\caption{Two-layer early fusion and late fusion ReLU networks trained on a linear task. The setting is the same as \cref{fig:dynamics_2L}, except that ReLU activation is added to the hidden layer. (a,b) Time trajectories of loss and total weights in the two-layer early fusion and
late fusion ReLU network.}
\label{fig:ReLU_2L}
\end{figure}

Two-layer ReLU networks are trained to learn the same linear task introduced in \cref{fig:dynamics_2L}. The loss and weights trajectories in \cref{fig:ReLU_2L} are qualitatively the same as \cref{fig:dynamics_2L}, except that learning evolves about two times slower and the converged total weights are two times larger.

\subsection{ReLU Networks with Nonlinear Task}

\begin{figure}[h]
\vspace{3ex}
\centering
\subfloat[Early fusion, $\sigma_\A=1$ \label{fig:xor-early-1var}]
{\centering
\includegraphics[width=0.23\linewidth]{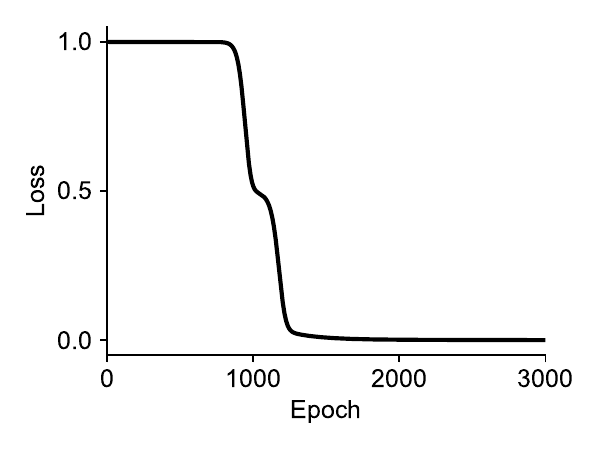}
\includegraphics[width=0.26\linewidth]{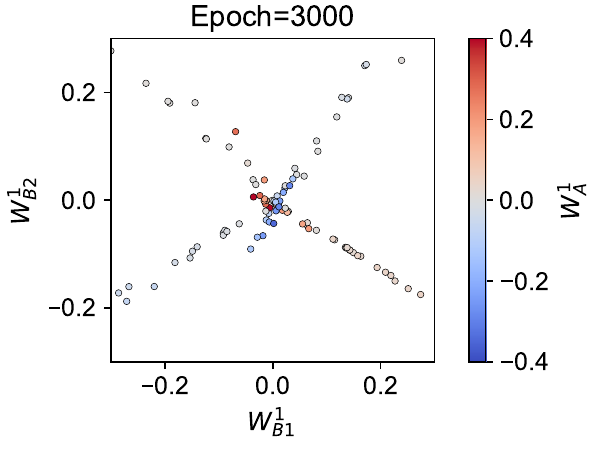}}
\hfill
\subfloat[Late fusion, $\sigma_\A=1$ \label{fig:xor-late-1var}]
{\centering
\includegraphics[width=0.23\linewidth]{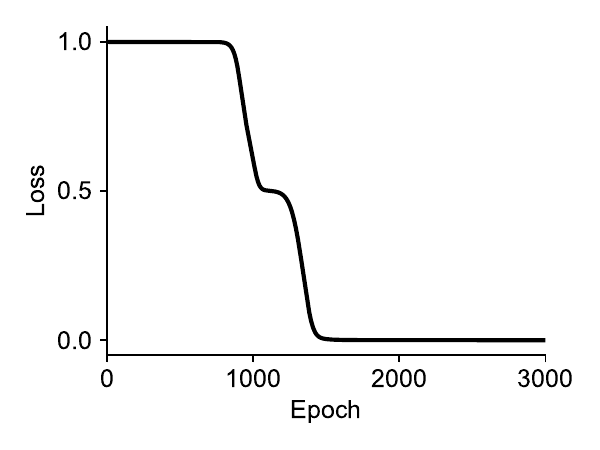}
\includegraphics[width=0.26\linewidth]{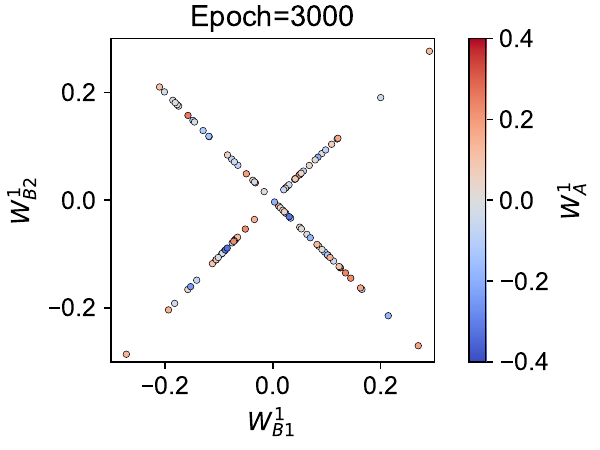}}
\\[2ex]
\subfloat[Early fusion, $\sigma_\A=2$ \label{fig:xor-early-2var}]
{\centering
\includegraphics[width=0.23\linewidth]{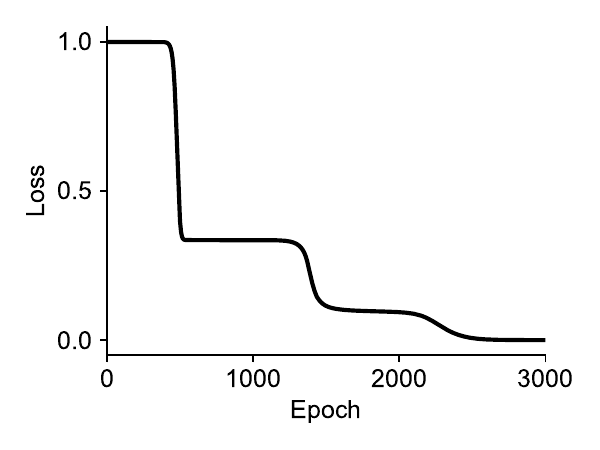}
\includegraphics[width=0.26\linewidth]{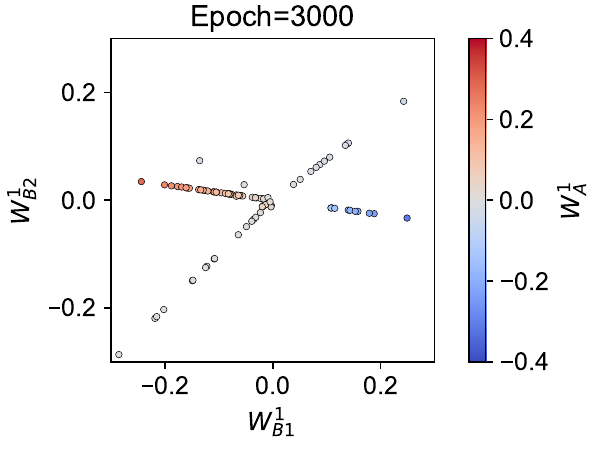}}
\hfill
\subfloat[Late fusion, $\sigma_\A=2$ \label{fig:xor-late-2var}]
{\centering
\includegraphics[width=0.23\linewidth]{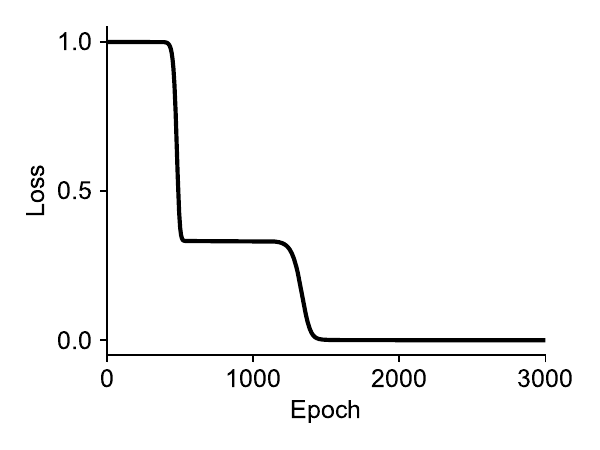}
\includegraphics[width=0.26\linewidth]{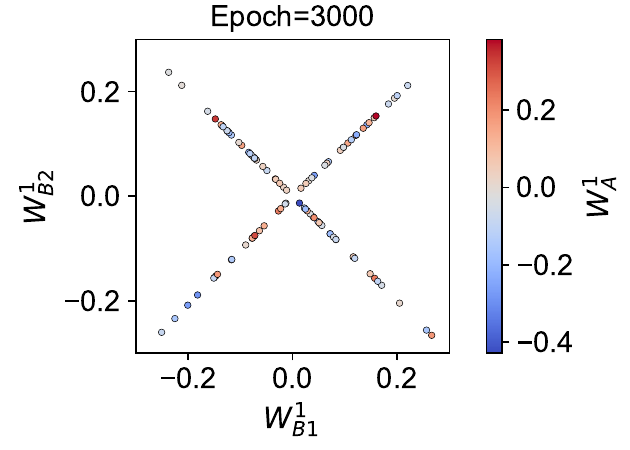}}
\\[2ex]
\subfloat[Early fusion, $\sigma_\A=3$ \label{fig:xor-early-3var}]
{\centering
\includegraphics[width=0.23\linewidth]{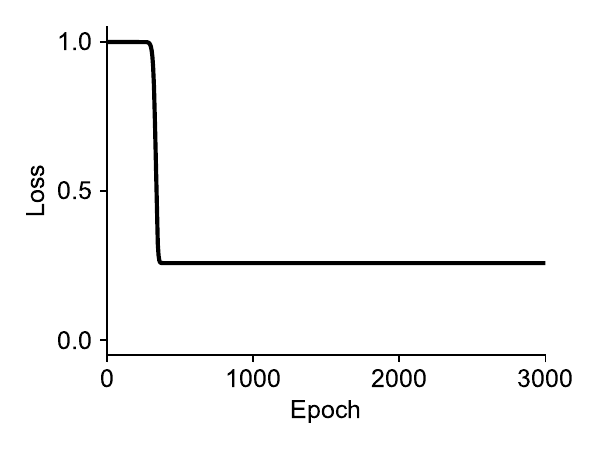}
\includegraphics[width=0.26\linewidth]{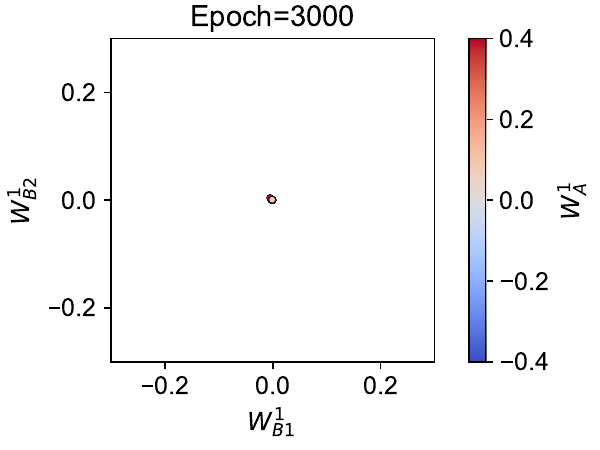}}
\hfill
\subfloat[Late fusion, $\sigma_\A=3$ \label{fig:xor-late-3var}]
{\centering
\includegraphics[width=0.23\linewidth]{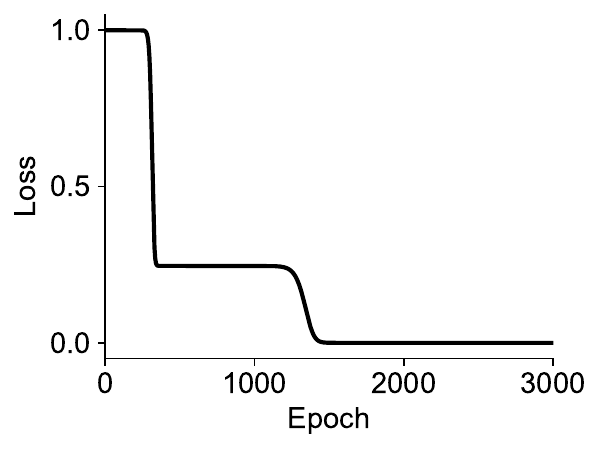}
\includegraphics[width=0.26\linewidth]{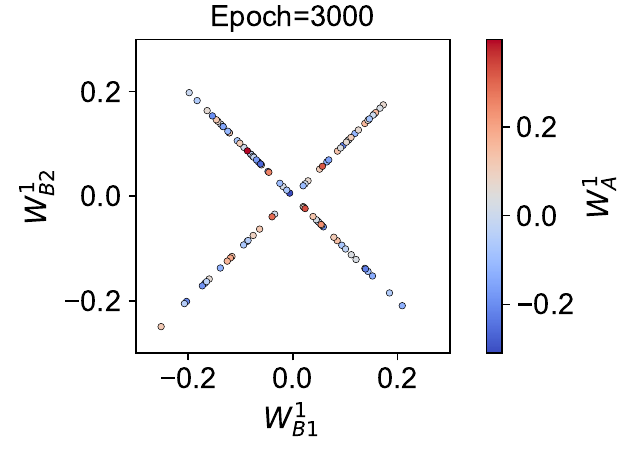}}
\caption{Two-layer early fusion and late fusion ReLU networks trained on an XOR and linear task. The early fusion network has 100 hidden neurons. The late fusion network has 100 hidden neurons in both branches. In every panel, the loss trajectory is plotted on the left and the first-layer weights at the end of training are plotted on the right. The first-layer weights are three-dimensional: the two dimensions of $\mW_\B^1$ are plotted as the position of the dots and $\mW_\A^1$ is plotted as the color of the dots.
(a,c,e) Two-layer early fusion ReLU networks. (b,d,f) Two-layer late fusion ReLU networks. Complementary videos can be found on our website.}
\label{fig:XOR-ReLU}
\end{figure}

We consider a simple nonlinear task of learning $y=\vx_\A + \text{XOR}(\vx_\B)$, where $\vx_\A \in \sR, \vx_\B \in \{ [1,1], [1,-1], [-1,1],\allowbreak [-1,-1] \}$. The term $\text{XOR}(\vx_\B)$ refers to performing exclusive-or to the two dimensions of $\vx_\B$. We train two-layer early fusion and late fusion ReLU networks to learn this task with variance of the linear modality $\Var (\vx_\A) \equiv \sigma_\A =1,2$ or $3$. We inspect the features in ReLU networks by plotting the first-layer weights since the features in a two-layer ReLU network lie in its first-layer weights \citep{xie17relu}.

As shown in \cref{fig:xor-late-1var,fig:xor-late-2var,fig:xor-late-3var}, two-layer late fusion ReLU networks always solve the task by consistently forming the four perpendicular XOR features. We can see two transitions in the loss trajectories of late fusion ReLU networks, which are similar to the transitions in linear networks. Late fusion ReLU newtorks learn the XOR modality during one transition stage and learn the linear modality during the other. As an example of converged features shown in \cref{fig:xor-late-2var}, $\mW_\B^1$ has taken on the rank-two XOR structure and $\mW_\A^1$ has grown in scale while preserving its independence from $\mW_\B^1$ at initialization.

As shown in \cref{fig:xor-early-1var,fig:xor-early-2var,fig:xor-early-3var}, two-layer early fusion ReLU networks struggle to extract the XOR features. In the early stage of training, features in $\mW^1$ favor particular directions in the three-dimensional space that are most correlated with the target output. In comparison with features in late fusion ReLU networks, the first layer weights for modality A do not preserve its independence from weights for modality B as shown in \cref{fig:xor-early-2var}.
The larger the variance of the linear modality, the closer the favorable direction is to the direction of the linear modality. In later stage of training, features in $\mW^1$ can rotate or scatter, giving rise to multiple transition stages as shown in \cref{fig:xor-early-2var}. For a large variance in the linear modality, the features are closely aligned with the linear modality direction and the network can be stuck in a local minimum, failing to learn from the XOR modality as in \cref{fig:xor-early-3var}.

\section{Noisy MNIST Classification}  \label{supp:mnist}
We perform experiments to validate that our theoretical results, derived for multimodal linear networks, can extend to more realistic scenarios. 
We train deep fully-connected ReLU networks and deep convolutional networks with different fusion layer depths on a noisy MNIST digit classification task.
The results qualitatively align with our conclusion that the deeper the layer at which fusion occurs, the longer it takes to learn from the weaker modality.

\begin{figure}[t]
\vspace{3ex}
\centering
\subfloat[Corrupt $\vx_\B$ (prob 0.6) \label{fig:mnist-noiseB}]
{\centering
\includegraphics[width=0.25\linewidth]{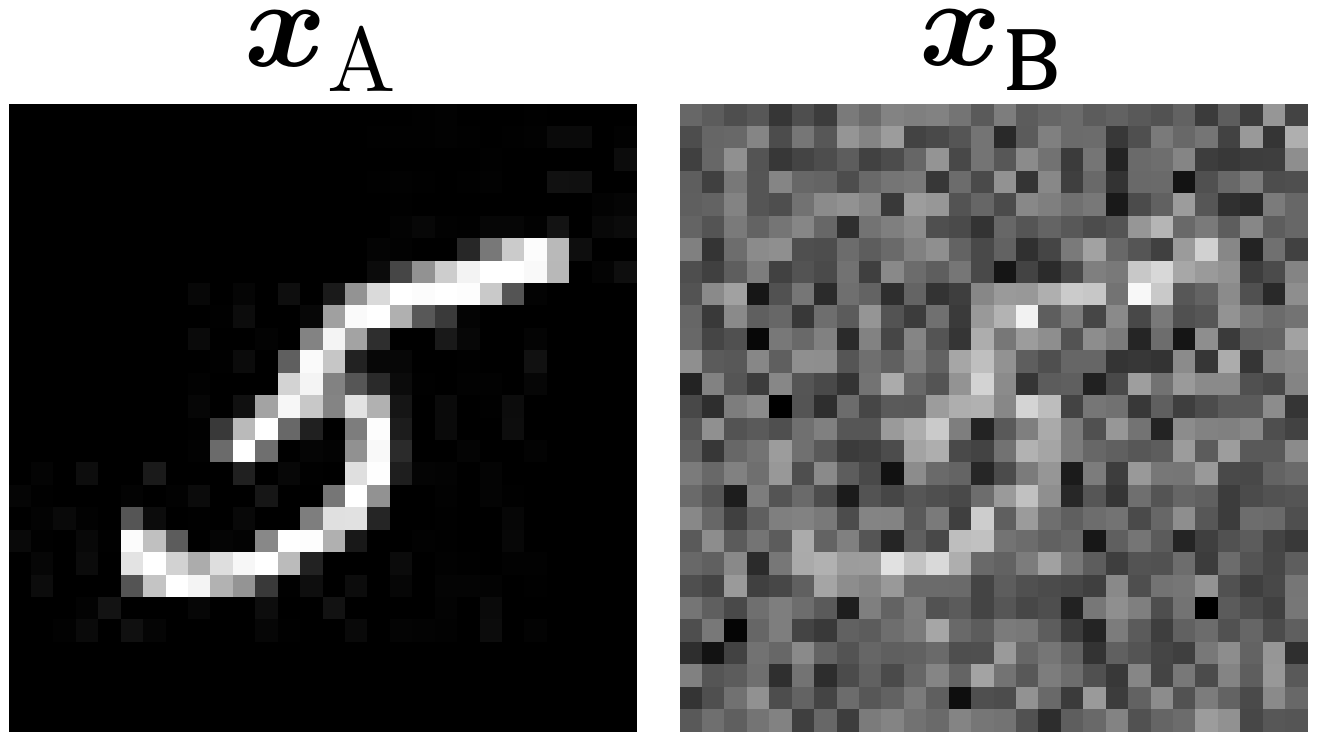}}
\hspace{4ex}
\subfloat[Corrupt $\vx_\A$ (prob 0.2) \label{fig:mnist-noiseA}]
{\centering
\includegraphics[width=0.25\linewidth]{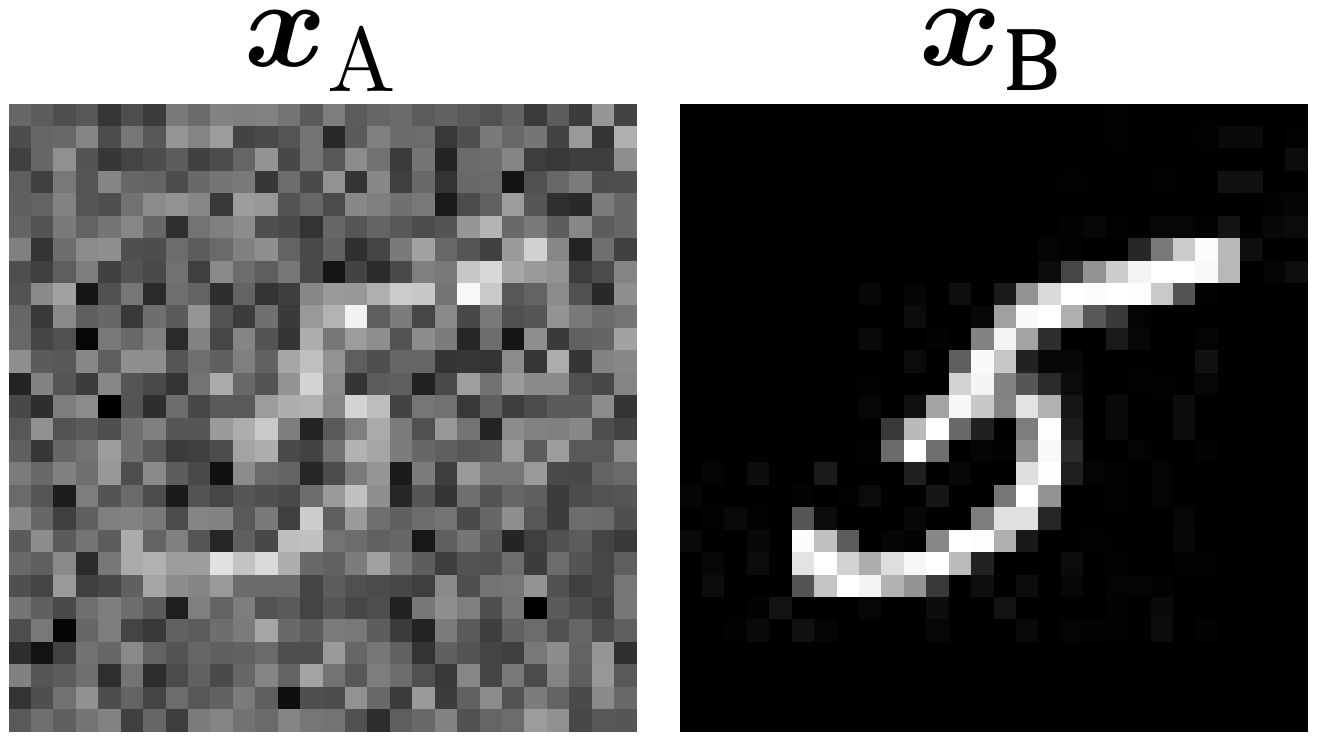}}
\hspace{4ex}
\subfloat[Uncorrupted (prob 0.2) \label{fig:mnist-nonoise}]
{\centering
\includegraphics[width=0.25\linewidth]{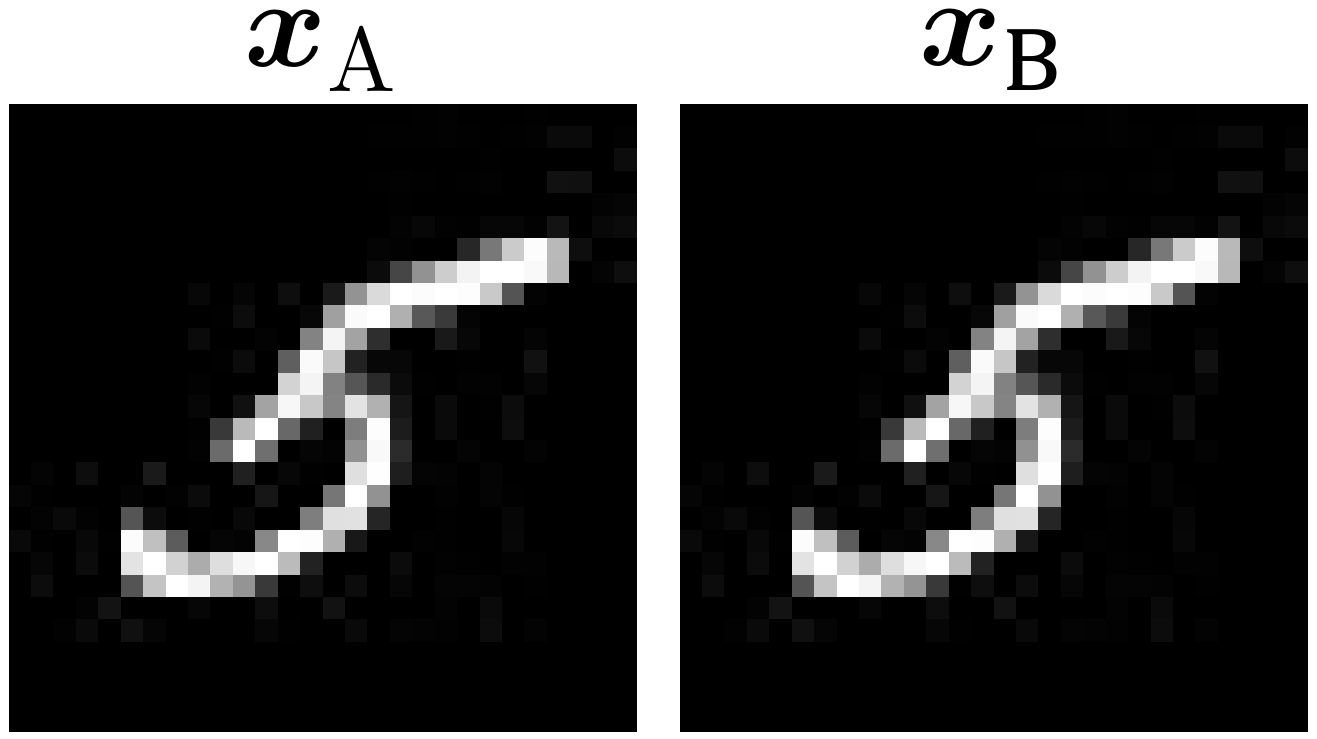}}
\caption{Noisy MNIST dataset.}
\label{fig:mnist-sample}
\end{figure}

\begin{figure}[t]
\centering
\subfloat[CNN $L_f=1$]
{\centering
\includegraphics[width=0.18\linewidth]{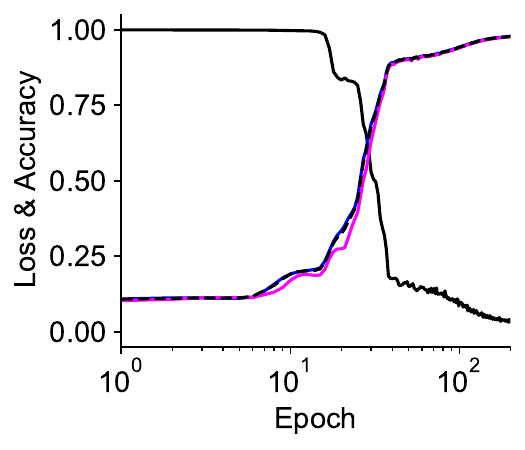}}
\subfloat[CNN $L_f=2$]
{\centering
\includegraphics[width=0.18\linewidth]{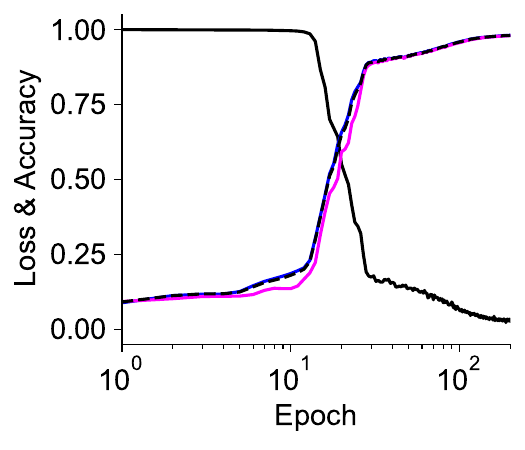}}
\subfloat[CNN $L_f=3$]
{\centering
\includegraphics[width=0.18\linewidth]{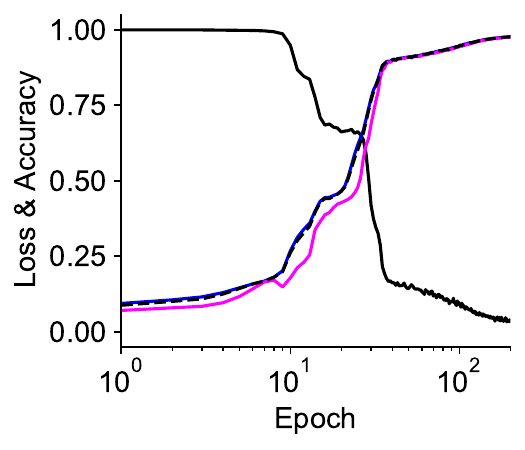}}
\subfloat[CNN $L_f=4$]
{\centering
\includegraphics[width=0.18\linewidth]{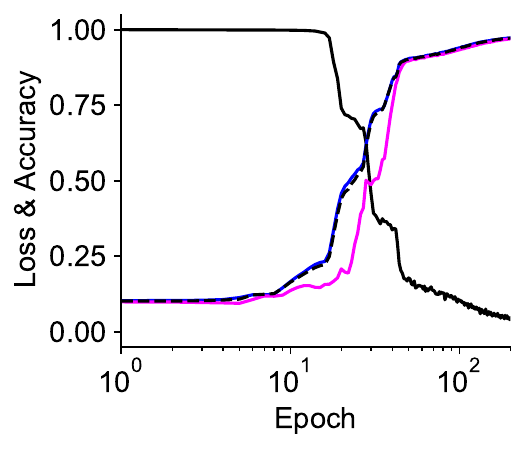}}
\subfloat[CNN $L_f=5$]
{\centering
\includegraphics[width=0.267\linewidth]{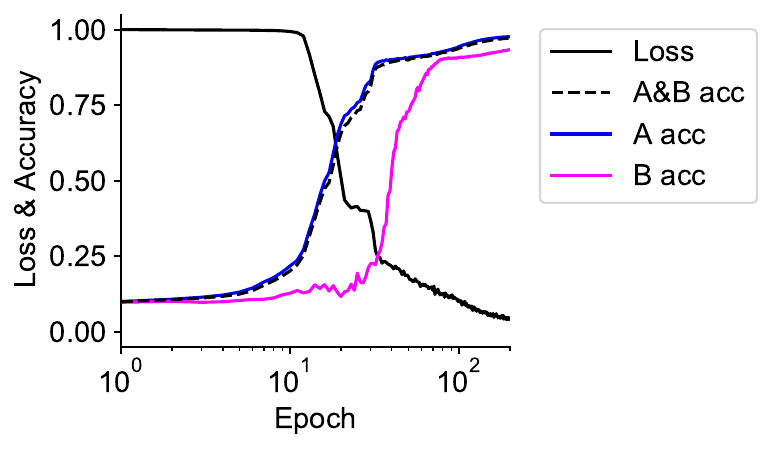}}
\caption{Loss and accuracy trajectories of multimodal convolutional networks. Six-layer ReLU newtorks with different fusion layer depths are trained on the noisy MNIST dataset shown in \cref{fig:mnist-sample}. The solid black curve plots the training loss. The dotted black curve plots the multimodal classification accuracy, for which both branches are presented with an uncorrupted testset image. The blue curve plots the unimodal classification accuracy of modality A, for which branch A is presented with an uncorrupted testset image and branch B with a blank image. Similarly, for modality B (purple curve). 
The trajectories are averaged over five random seeds.}
\label{fig:cnn-mnist}
\end{figure}

In the noisy MNIST classification, the multimodal network receives two written digit images as two modalities, one of which may be corrupted by Gaussian noise. With probability 0.6, modality B is corrupted as in \cref{fig:mnist-noiseB}. With probability 0.2, modality A is corrupted as in \cref{fig:mnist-noiseA}. With probability 0.2, both modalities are uncorrupted as in \cref{fig:mnist-nonoise}.
This is a common scenario in multimodal learning where the dominating modality varies per sample.

The deep fully-connected ReLU networks and deep convolutional networks are trained with SGD with cross-entropy loss on the noisy MNIST dataset. The batch size is 1000.
The learning rate at the beginning of training is $0.04$ for the fully-connected ReLU networks and $0.002$ for the convolutional networks. We use a learning rate scheduler that decays the learning rate by $0.996$ every epoch. Pytorch's default initialization is used.

The fully-connected ReLU networks have depth $L=5$ and fusion layer $L_f=1,2,\cdots,5$. The architecture is the same as multimodal deep linear networks except for the ReLU activation.
The network width is 500 except for the first layer. 

The convolutional networks have six layers: the first five layers are convolutional layers with ReLU activation and the last layer is a fully-connected layer. At a convolutional fusion layer, two inputs are concatenated along the channel and then passed to post-fusion layers. The number of output channels is 32; the kernal size is 3; the stride is 1.

Similar to what we have done with multimodal linear networks, we record the time ratio for the noisy MNIST experiments.
The time ratio is computed by dividing the time when the unimodal accuracy of the firstly learned modality reaches $50\%$ by that of the second modality.
The time ratio serves as a metric for the relative speed at which the network learns from the second modality compared to the first.
All experiments are repeated five times from five random seeds. We report the average and the standard deviation in \cref{table}.

\section{Implementation Details \label{supp:implementation}}
We provide our code at \href{https://github.com/yedizhang/unimodal-bias}{https://github.com/yedizhang/unimodal-bias}.

\subsection{Multimodal Linear Networks}
Early fusion networks have 100 neurons in every layer. Late fusion networks have 100 neurons in every layer for both branches. Intermediate fusion networks have 100 neurons in every pre-fusion layer for both branches and 100 neurons in every post-fusion layer. 
All networks are trained with full batch gradient descent with learning rate $0.04$. Thus the time constant $\tau$, which appeared in many differential equations such as \cref{eq:dWAdWB,eq:dW}, is $\tau=1/0.04=25$ for our experiments. 

For \cref{fig:timesweep_2L,fig:amountsweep_2L,fig:rho-sweep-4L,fig:ratio-sweep-4L,fig:init-sweep-4L,fig:fcn-mnist,fig:cnn-mnist,table}, we run experiments with five different random seeds and report the average.

\subsection{Data Generation}
For data generation, we sample input data points from zero mean normal distribution $\vx \sim \mathcal N (\vzero, \mSigma)$ and generate the corresponding target output from a groundtruth linear map $y = \vw^* \vx$. All datasets do not contain noise except \cref{fig:Eg}. The number of training samples is 8192 for all experiments except \cref{fig:Eg}. Hence, all experiments, except \cref{fig:Eg}, fall into the underparameterized regime where the training loss well reflects the generalization error.

Note that we do not lose generality by using linear datasets because the learning dyanmics of linear networks as given in \cref{eq:dWAdWB,eq:dW} only concern the input correlation matrix $\mSigma$ and input-output correlation matrix $\mSigma_{y\vx}$. Hence, datasets generated from any distribution and target map with the same $\mSigma, \mSigma_{y\vx}$ will have the same learning dynamics in linear networks.

\subsection{Specifications for Each Figure}

\textbf{\cref{fig:dynamics_2L}}.
Inputs $\vx_\A,\vx_\B$ are scalars with covariance matrix $\mSigma=\text{diag}(4,1)$. The target output is generated as $y = \vx_\A + \vx_\B$. All weights in the early fusion network and the late fusion network are initialized with independent random samples from $\mathcal N(0,10^{-9})$.

\textbf{\cref{fig:sweep_2L}}.
Inputs $\vx_\A,\vx_\B$ are scalars with different covariance matrices parameterized as $\mSigma = [\sigma_\A^2, \rho \sigma_\A \sigma_\B; \rho \sigma_\A \sigma_\B, \sigma_\B^2]$, that is $\sigma_\A,\sigma_\B$ are the variances of modality A,B and $\rho$ is their correlation coefficient. The target output is generated as $y=\vx_\A+\vx_\B$. All weights are initialized with independent random samples from $\mathcal N(0,10^{-9})$.
In this two-dimensional case, the derived time ratio reduces to $1+(\frac{\sigma_\A^2}{\sigma_\B^2}-1) / (1-\rho^2)$; the amount of mis-attribution reduces to $\rho \frac{\sigma_\B}{\sigma_\A}$. 

\textbf{\cref{fig:superficial}}.
In panel (a),  the input covariance matrix is $\mSigma=\text{diag}(9,1)$ and the target output is $y=\vx_\A+4\vx_\B$. The dotted gray line is below the dotted black, meaning modality B contributes more to the output. The prioritized modality is therefore not the modality that contributes more to the output. In panel (b), the input covariance matrix is $\mSigma=\text{diag}(16,1)$ and the target output is $y=\vx_\A+3\vx_\B$. The dotted black line is below the dotted gray, meaning modality A contributes more to the output. The prioritized modality is the modality that contributes more to the output. All weights are initialized with independent random samples from $\mathcal N(0,10^{-9})$.

\textbf{\cref{fig:Eg}}.
The inputs $\vx_\A, \vx_\B \in \sR^{50}$ have covariance $\mSigma_\A=\mI, \mSigma_\B=3\mI, \mSigma_{\A\B}=\vzero$. The target output is generated as $y = \vw^* \vx + \epsilon$, where $\vw^*=\vone/10$ and the noise is independently sampled from $\epsilon \sim \mathcal N(0,0.5^2)$. All weights are initialized with independent random samples from $\mathcal N(0,10^{-9})$.

\textbf{\cref{fig:depth4_sweep}}.
Inputs $\vx_\A,\vx_\B$ are scalars with different covariance matrices parameterized as $\mSigma = [\sigma_\A^2, \rho \sigma_\A \sigma_\B; \rho \sigma_\A \sigma_\B, \sigma_\B^2]$. The target output is generated as $y=\vx_\A+\vx_\B$. Except the initialization scale sweep in \cref{fig:init-sweep-4L}, pre-fusion layer weights are initialized with independent random samples from $\mathcal N(0,10^{-4})$; post-fusion layer weights are initialized with independent random samples from $\mathcal N(0,2\times 10^{-4})$.

\end{document}